\documentclass{article}

\usepackage{PRIMEarxiv}

\usepackage[utf8]{inputenc} % allow utf-8 input
\usepackage[T1]{fontenc}    % use 8-bit T1 fonts
\usepackage{hyperref}       % hyperlinks
\usepackage{url}            % simple URL typesetting
\usepackage{booktabs}       % professional-quality tables
\usepackage{amsfonts}       % blackboard math symbols
\usepackage{nicefrac}       % compact symbols for 1/2, etc.
\usepackage{microtype}      % microtypography
\usepackage{fancyhdr}       % header
\usepackage{graphicx}       % graphics
\graphicspath{{media/}}     % organize your images and other figures under media/ folder
\usepackage{amsmath}
\usepackage{multirow}
\usepackage{xcolor}
\usepackage{soul}
\usepackage{bm}
\usepackage{CJKutf8}  % todo: check
\usepackage{array}
\usepackage{natbib}

\title{Defending Against Authorship Identification Attacks
}

\author{
  Haining Wang \\
  Indiana University Bloomington \\
  Bloomington, Indiana, USA\\
  \texttt{hw56@indiana.edu} \\
}

\begin{document}
\maketitle

\begin{abstract}
Authorship identification has proven unsettlingly effective in inferring the identity of the author of an unsigned document, even when sensitive personal information has been carefully omitted.
In the digital era, individuals leave a lasting digital footprint through their written content, whether it is posted on social media, stored on their employer's computers, or located elsewhere. 
When individuals need to communicate publicly yet wish to remain anonymous, there is little available to protect them from unwanted authorship identification.
This unprecedented threat to privacy is evident in scenarios such as whistle-blowing.
Proposed defenses against authorship identification attacks primarily aim to obfuscate one's writing style, thereby making it unlinkable to their pre-existing writing, while concurrently preserving the original meaning and grammatical integrity.
The presented work offers a comprehensive review of the advancements in this research area spanning over the past two decades and beyond.
It emphasizes the methodological frameworks of modification and generation-based strategies devised to evade authorship identification attacks, highlighting joint efforts from the differential privacy community.
Limitations of current research are discussed, with a spotlight on open challenges and potential research avenues.
\end{abstract}

% keywords can be removed

\section{Introduction\label{sec: introduction}}

\subsection{Fingerprinting Writers}

Writing style enables more nuanced communication, extending beyond mere semantics to convey aspects including attitude, humor, formality, and politeness \citep{ohmann1964generative}.
However, the style present in a person's writing can potentially be used to identify them through the use of authorship fingerprinting techniques, namely \textit{stylometry}.
Research has shown that stylometric analysis can accurately identify the author of a text with unknown authorship, achieving an accuracy rate higher than what would be expected by chance \citep{holmes1998evolution, juola2008authorship, koppel2009computational, stamatatos2009survey}.
For example, an authorship identification model with several hundred hand-crafted linguistic features only needs several thousand English words \citep{eder2015does, rao2000can} to achieve over 90\% accuracy when given a set of 50 candidate authors \citep{abbasi2008writeprints} and 25\% accuracy given 100,000 candidates \citep{narayanan2012feasibility}.
Deep neural network-based models is capable of fingerprinting writers on a very large scale \citep{fabien2020bertaa, hu2020deepstyle}.
Making predictions using training data from a distinct domain has proven feasible \citep{overdorf2016blogs, barlas2020cross}.

The increasing availability of written content makes authorship fingerprinting susceptible to abuse.
In today's digital age, people routinely compose, exchange, share, and store their writing online. 
Approximately 575 thousand tweets are posted, and 12 million iMessages are sent every minute \citep{domo2020data}. 
A significant portion of this data contains information that can be used to identify individuals, such as writings associated with their identities or other demographic attributes.
According to a survey conducted by the Pew Research Center, 38\% of 792 American adults reported that ``things they have written using their names'' were available online \citep{rainie2013anonymity}.
Also, a substantial amount of personal writing is stored on employers' computers, often in the form of memoranda, emails, and work reports.
These writings, associated with an individual's identity and potentially found on social media or on an employer's device, may be used as a reference against one's anonymization efforts.

From a communication security perspective, efforts to remain anonymous, such as using Virtual Private Networks (VPNs) or disposable email addresses, may not provide sufficient protection against authorship fingerprinting techniques, which rely solely on the end message and are indifferent to the communication channel.
Even worse, individual writing can be combined with pieces of background information to further compromise privacy.
This concern is underscored by the fact that, using only gender, birth date, and post code, 63\% of the U.S. population could be uniquely identified with census data from the year 2000 \citep{golle2006revisiting}.
Effective authorship fingerprinting could exacerbate these privacy risks.

In summary, authorship fingerprinting presents a formidable challenge for individuals seeking to remain anonymous while sharing even modest amounts of written content.

\subsection{Defenses Against Authorship Fingerprinting}
Since 2000, various countermeasures have been emerged to defend against stylometric analysis.
These strategies aim to modify one's style found in a new document so that it is not linkable to its original author while ensuring successful communication.
Throughout the survey, we will use ``adversarial stylometry'' and ``defense against authorship identification attacks'' interchangeably, referring to efforts that frustrate stylometric analysis during the test phase.
As an example, we will next demonstrate how an author's identity can be vulnerable to stylometric analysis and introduce one of its countermeasures.

\paragraph{Machine Zone, Glassdoor, and John Doe}

An anonymous review entitled ``A Scandal!'' critically evaluated Machine Zone, Inc, a mobile game company from Silicon Valley.\footnote{Parts of this example were adopted from \citet{kosseff2022anonymous}.}
The review mentions, ``In July 2014, [Machine Zone's] CEO announced that they raised \$250,000,000 (250 million) from JP Morgan, based on a total value 3 billion dollars  ([sic]). After one year has passed, this hasn't been verified by any other resources ([sic]). The CEO has never mentioned it again.''
The review also complains about unbalanced workloads among teams and poor strategic planning. 

The review was posted on Glassdoor.com, a platform for employees to rate their past and current employers based on their work experiences.
Machine Zone sued the anonymous poster, referred to as John Doe, for breach of contract, claiming that Doe ``disclos[ed] to third parties Machine Zone's confidential, non-public information.'' 
In an attempt to uncover the identity of John Doe, Machine Zone filed a subpoena requesting that Glassdoor provide author-identifying information.
Glassdoor refused to cooperate, stating that disclosure of the poster's identity would violate their First Amendment right to speak anonymously. 
A trial court granted a motion filed by Machine Zone to compel Glassdoor to assist with the discovery, but the appellate court reversed the motion, stating that as a website operator, Glassdoor ``may assert the First Amendment rights of their anonymous contributors.''

Regardless of the review's credibility, Glassdoor's protection of the author's anonymity is a victory for freedom of speech. 
However, in hindsight, Doe was lucky insofar as Machine Zone did not employ linguistic forensics, as this technique can reveal the identity of an anonymous author by analyzing the writing style, even if their IP or email address is protected by Glassdoor.

\paragraph{What If Machine Zone Applies Stylometric Analysis?}

Let's return to the very beginning of the Machine Zone v. John Doe case, and assume that before launching the lawsuit, the company happened to learn that a stylometric analysis could reveal John Doe's identity. 
First, the review is publicly available and long enough to use as the test sample, with its author being the target of the discovery.
Second, it appears that the review has not been consciously edited and thus contains enough stylistic clues, such as the omission of ``of'' in ``a total value 3 billion dollars.''
Third, it would be straightforward to assemble the candidate pool, and Machine Zone would have had access to a wealth of employee documents, including work reports, memoranda, and emails. These could serve as training material for a stylistic discriminator.

For illustration, we approximate the stylometric analysis using a subset of the Riddell-Juola corpus v.~1.0 \citep{wang2022reproduction}.\footnote{Note that for simplicity, we only use the training split of the control group of the Riddell-Juola corpus.}
The corpus includes 17 authors, 7 of whom are male and 10 are female, reflecting the imaginary candidate pool formed by Machine Zone. 
Each candidate is associated with approximately 6,500 words of pre-existing writing, with all but one of these samples used for training, mirroring the work documents that the company would have access to. 
The withheld sample is presumably written by Doe and is used as the test sample, parallel to an anonymously posted review on Glassdoor.

To identify John Doe from the candidate pool, the company can simply use a multi-class linear support vector machine (SVM) classifier with frequencies of function words (i.e., words carrying little to no semantics but grammatically needed, such as ``is'' and ``that'') calculated from training samples.\footnote{Specifically, we use a standard SVM with an $\ell_2$ penalty and a regularization parameter set to $1.0$. Before fitting, the feature vectors are normalized by the length of each training sample and standardized by subtracting the mean and dividing by the standard deviation of the training samples. The function word list is adopted from \citet{koppel2009computational}.}
It turns out that the classifier can easily pick out the true author: John Doe was ranked first among the suspects with a probability of approximately 0.18, far better than a random guess (1/17 or approximately 0.06).

\begin{figure}
  \centering
  \includegraphics[width=\linewidth]{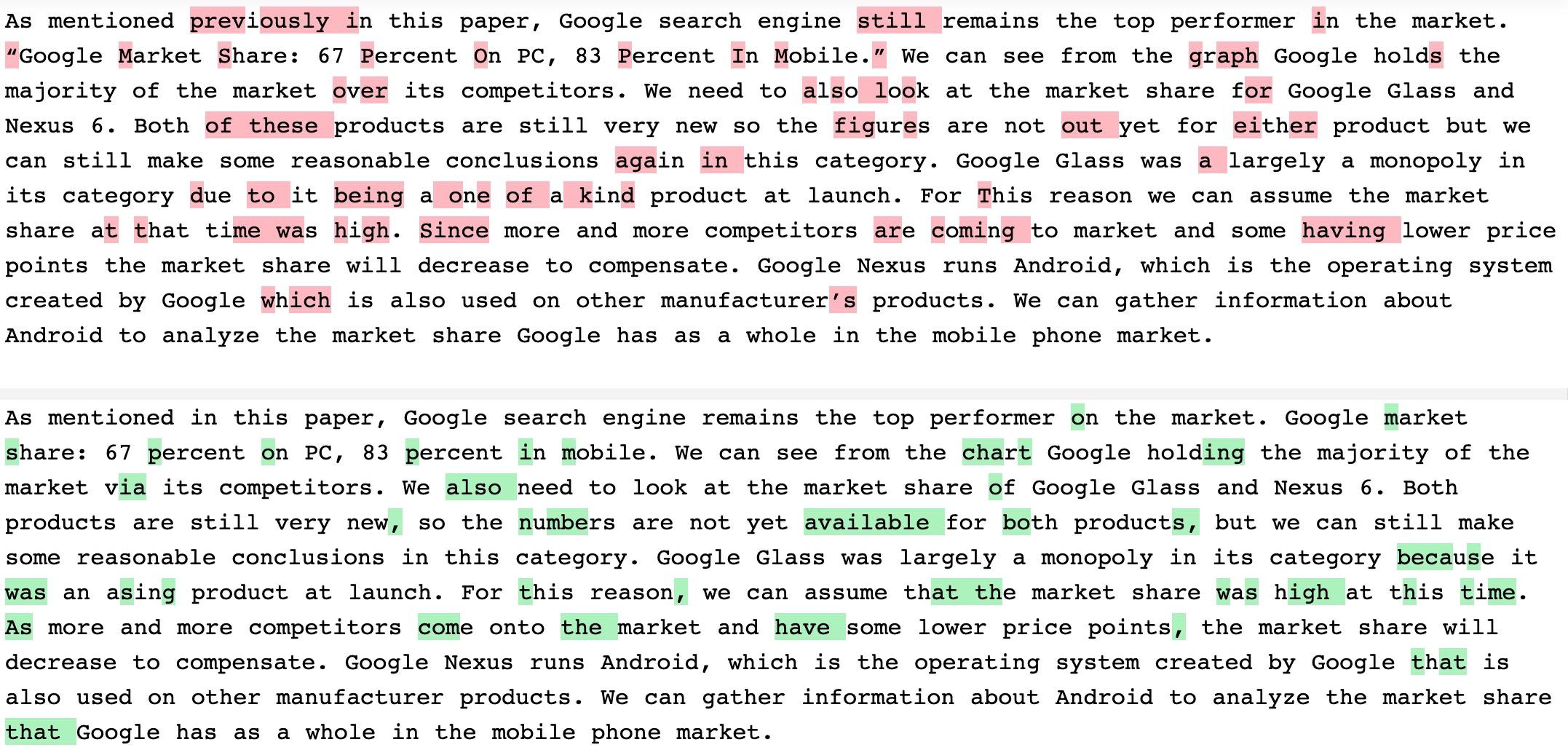}
  \caption{An illustration of how adversarial stylometry works in practice. The top panel presents a pre-existing writing sample chosen as the ``anonymous post.'' The lower panel is the back-translated version, where the sample is first translated to German, and then back to English. Differences are highlighted. Despite the meanings of the two writing samples being equivalent, a standard authorship attribution model is deceived into predicting the back-translated version is from a different author. Visualization sourced from \url{https://text-diff.com/}.}
  \label{fig: example}
\end{figure}

However, let's further assume that John Doe was aware of the potential for identification through his writing style. 
Before posting the review, Doe translated the post into German, then back into English, using translation software on his personal computer.\footnote{We use translateLocally \citep{bogoychev2021translatelocally}, a translator designed to work without an internet connection. Specifically, we use the ``base'' models for the round-trip translation.}
This time, when fitted with the same training data, the model falsely believes the writing is from someone else and ranks Doe in third place.
The simple trick of round-trip translation only changes a handful of words but makes a significant difference for John Doe.
The back-translated sample is compared to the ``anonymous'' post in Figure~\ref{fig: example}.

This fingerprinting scenario is purely fictional, it provides an illustration of stylometric analysis and adversarial stylometry in action. 
Although basic adversarial tactics, like the back-translation strategy used in our example, can somewhat obscure an author's style, it is clear that maintaining anonymity against sophisticated authorship fingerprinting attacks demands more advanced strategies.

\subsection{Implications\label{sec: implications}}

Anonymity is integral to free speech and privacy.
Traditional research has emphasized the security of the communication channel, rather than the content, in considering how to preserve one's anonymity.
However, with the advancement of authorship fingerprinting techniques, extra caution must be exercised in the content shared, as an individual's writing style can be analyzed to reveal their identity even among a large group of candidates.
Equipped with advanced technology and an abundance of readily available personal identification information, governments or well-resourced organizations can stealthily pinpoint individuals through stylometric analysis, often bypassing oversight or checks and balances.
This threat is particularly pressing for whistleblowers, activists, and journalists, whose messages are intended for the general public.
Therefore, the development of adversarial stylometry has unique and significant societal implications, as it enables individuals to speak anonymously by complementing traditional channel protection approaches.
Advances in this field are expected to foster free speech by alleviating the impact of chilling effects and enhancing privacy.

Despite the growing interests in using machine learning to enhance anonymity and privacy across various data types, including personal data stored in a table \citep{abadi2016deep}, images \citep{fan2019practical}, speech \citep{fang2019speaker}, and video \citep{maximov2020ciagan}, no systematic review for anonymizing personal writings exists to date.
The topic discussed is closely related to style transfer \citep{jin2022deep} and text revision \citep{li2022text}.

While our defense provides a means to enhance anonymity, we must acknowledge that no anonymity-enhancing techniques are impervious to ill-intent, such as cyberbullying, harassment, or even illegal activities.
Nevertheless, we maintain that these concerns should not be overstated.
Stylometry is a broad field, with a significant portion dedicated to studies that could serve forensic objectives.
This inherent dichotomy could foster a balance between the necessity of anonymity protection and the imperative to address unethical or criminal actions cloaked in anonymity.
% Furthermore, when appropriately adjusted, the perturbed samples produced by adversarial stylometry could be used to augment training data, thereby aiding in the creation of more robust stylometric models.

\subsection{Scope\label{sec: scope}}
This survey concerns circumventing stylometric analysis at test time.
A common scenario is that an individual prefers to keep their identity hidden when sending a message consisting of at least several sentences.
We can reasonably assume that:
\begin{enumerate}
    \item Individuals seeking to evade authorship fingerprinting would already redact personally identifiable information (PII) from the document to be obfuscated; and
    \item A defense operates without any control over the adversary's data, model, or training process.\footnote{See \citet{wang2022upton} for a study interested in \emph{poisoning} adversary's training data.} Natural language processing-based approaches usually do not assume prior knowledge beyond training samples. However, this survey includes a review of privacy-preserving data publishing (PPDP) research. The goal of PPDP is to introduce noise into personal long documents to deter linkage attacks, namely stylometric analysis. PPDP-based approaches assume complete knowledge of the author, except for their identity. See discussions in Sections~\ref{sec: dp} and~\ref{sec: taxonomy_available_knowledge}. 
\end{enumerate}

The survey focuses on studies that address the message content \textit{per se}, leaving the security of the communication channel and penmanship-based forensics to the scrutiny of other communities.
Encryption and steganography, while useful for maintaining anonymity, are not reviewed as they are not suitable for communicating with general audiences.
Additionally, obscuring artificial languages (e.g., source code), obvious PII, and documents generated from large language models (LLMs) also fall outside the scope of this survey.

In light of the evolving landscape of stylometric analysis, this review offers a fresh perspective, distinguishing itself from prior works in several key areas.
\citet{zhang2020adversarial} reviewed how strategically corrupted samples influence the decisions of sentiment and topic classifiers.
\citet{potthast2016author}, \citet{neal2017surveying}, and \citet{grondahl2019text} each include sections on early attempts at adversarial stylometry.
\citet{zhao2022survey} conducted a review of noising long text from a differential privacy perspective, with a particular emphasis on token-level privacy mechanisms.
We have conducted an up-to-date, systematic review of defenses against stylometric analysis.
% We also dedicate a significant portion of the text to provide an overview of stylistic features and their representations.
% Compared to earlier reviews on stylometry \citep{juola2008authorship, stamatatos2009survey, koppel2009computational, neal2017surveying}, we place greater emphasis on stylistic features and their use in cross-domain generalization, some of which reflects advancements in deep neural network-based models.
This review assumes that the audience is comfortable with machine learning and natural language processing (NLP), but may not be familiar with stylometric studies.

\paragraph{Road Map}
The rest of the paper is organized as follows.
% todo: reference
We begin by reviewing authorship identification tasks, the counterparts of defenses, and personal writing style that reveal identity in Section~\ref{sec: stylometry_style}.
In Section~\ref{sec: adversarial_stylometry}, we formally introduce adversarial stylometry, including its brief history, formalization, taxonomy, and evaluation.
Next, we review the architecture, objectives, and performance of manual (Section~\ref{sec: manual}), human-machine collaborative (Section~\ref{sec: human_computer_collaboration}), and automatic (Section~\ref{sec: automatic}) approaches.
The survey concludes with a discussion on the current theoretical underpinnings of adversarial stylometry, the open challenges in the field, and the future prospects that could address these challenges or open new avenues for research and application.

\section{Stylometry \& Style}\label{sec: stylometry_style}

Stylometry can be used to uncover the identity, demographic characteristics, and other personal information of an author through the analysis of their writing style.
The field can be divided into subtasks based on the specific personal attributes they aim to uncover.
The primary \emph{adversary} considered in this survey is the individual or entity performing \emph{authorship identification}, which involves determining the identity of an author.
Studies that aim to \emph{profile} other personal aspects of an author will be discussed when relevant.\footnote{\emph{Authorship profiling} is to determine demographic attributes (e.g., gender and age), psychometric attributes (e.g., personality type), and immediate privacy attributes (e.g., mother tone and education) from one's writing.
While it does not directly identify the author's identity, it can be used to narrow down a candidate pool, possibly sharply.}

\subsection{Authorship Identification}

\subsubsection{Authorship Attribution}
Authorship identification infers the identity of an author from a text of unknown authorship.
It is typically modeled as a multi-class, single-label classification problem: given a closed set of candidates, the goal is to determine the true author (i.e., label) of the document.
The standard practice for identification is called \emph{authorship attribution}.

\subsubsection{Authorship Verification}
While the assumption of choosing one candidate from a closed set is often valid, it can also be a strong assumption that is difficult or impossible to meet in certain scenarios, such as for historical documents.
To address this, one option is to train the model to output ``none of the alternatives'' \citep{narayanan2012feasibility} or abstain from classification when the model's confidence is low \citep{noecker2012distractorless, xie2022many}.

An alternative approach to tackle open-set problems is \emph{authorship verification}, which transforms the multi-class classification problem into multiple binary classification problems: judging whether a specific author is the true author \citep{koppel2012fundamental}.
Verification may be a preferable option when only a portion of the candidate universe is known in advance.

\subsubsection{Premises}

There are numerous authorship issues that may be of interest; however, not all of them can be addressed via authorship identification.
The feasibility depends on whether the document under investigation meets the following premises.

\begin{itemize}
    \item The text of interest and pre-existing writing samples should be single-authored.\footnote{Although ascribing authorship to collaborative text is possible \citep{kestemont2015collaborative, xie2022many}, the field is understudied, perhaps because it is challenging to disentangle authorial components \citep{koppel2011unsupervised}.}
    \item Adequate reference writing samples are available. For the English language, the pre-existing documents for reference from each candidate should amount to several thousand words to be statistically informative. The text under investigation often consists of no fewer than several hundred words.
    \item The document whose authorship is unknown should be roughly aligned with the pre-existing samples in terms of factors known to influence writing style, such as genre \citep{baayen1996outside}, register \citep{goldstein2009person, stamatatos2013robustness, wang2021cross}, and input conditions \citep{overdorf2016blogs, wang2021mode}.
    \item The text under investigation has not been thoroughly edited or revised to reflect one's authentic writing style. Interference from an editor may hinder the natural flow of the author's writing style.
\end{itemize}

\subsubsection{Formalization}
A stylometric inquiry is customarily modeled as a classification problem.
Let the feature and label spaces be denoted as $X$ and $Y$, respectively.
Training data is noted as $D = \{\bm{x}_i, y_i\}_{i=1}^{n}$, where $n$ is the number of training samples.
During training, the classifier $f_{\bm{\theta}}$ seeks to optimize parameters $\bm{\theta}$ such that the loss $\ell$ between the true labels $y_i$ and the predicted labels $f_{\bm{\theta}}(\bm{x}_i)$ is minimized

\begin{equation}\label{eq: stylometry_as_classification}
  \arg \min_{\bm{\theta}} \sum_{D}  \ell(f_{\theta}(\bm{x}_i), \ y_i)
\end{equation}

For an attribution problem, $y_i$ corresponds to the identity of an author; in verification setups, $y_i$ is binary labels indicating whether two texts were written by the same author.
$X$ is the features or representations extracted from writing samples found in the training set, which stylometric tasks rely on to discriminate style.

\subsection{Writing Style}

% style is distinguishable
Studies in stylometry have demonstrated that individuals can be distinguished based on their use of language. 
A writer has a great degree of flexibility in their choice of words, sentence structure, and rhetoric when conveying the same meaning.
For example, the following sentences are virtually semantically equivalent.\footnote{Examples are taken from \citet{hoover1999language}.}
\begin{itemize}
    \item We were at a loss to find a suitable attendant for her.
    \item We were unable to find an appropriate attendant for her.
    \item We could not find the right care-giver for her.
    \item No one fitting could be found to tend her needs.
    \item Finding her a satisfactory attendant had us in a predicament.
    \item We were at our wits' end trying to find an appropriate attendant for her.
\end{itemize}

Despite the vast number of alternatives that exist, an author prefers certain expressions over others.
Over time, a writer's active vocabulary, preferred grammatical structure, and essay layout combine to create their distinctive and consistent \emph{writing style}.
In this survey, we refer to ``writing style'' as the hypothetical, authentic style possessed by an individual.\footnote{Researchers also use \emph{idolect} and \emph{stylome} to refer to an individual's writing style. We do not distinguish between these concepts and stick to ``writing style'' for simplicity.}
One's writing style can only be depicted using their complete body of work, analogous to the concept of ``population mean'' in statistics. 
The terms ``stylometric profile'' and ``stylistic representation'' describe an approximated representation of the individual's style derived from a portion of their previous writings, similar to a ``sample mean.''

% evidence supporting one's style is somehow unique
The formation of writing style is still a topic of debate \citep{johnstone1996linguistic, rudman2000non, love2002attributing, grieve2023register}\footnote{Please refer to \citet{ohmann1964generative} for a clear review of writing style.}, but the consensus is that individuals write differently, both in controlled experiments and in large corpora. 
In a field study, \citet{baayen2002experiment} attributed writings from eight Dutch students with similar educational backgrounds. 
In the experiment, each student was instructed to write nine prose pieces on fixed topics, with three topics from three different genres (i.e., fiction, argument, and description). 
Despite the strict control of the topics, genres, and educational backgrounds, \citet{baayen2002experiment} achieved over 80\% accuracy for pairwise attribution using leave-one-out cross-validation. 
The findings suggest that pre-existing writings are strongly associated with one's identity, as evidenced by the successful differentiation of authorship even when the topics of the held-out samples were unknown to the model during training. 
Additionally, with corpora consisting of tens of thousands of authors, \citet{zhu2021idiosyncratic} found that style representations learned with sentence-BERT with content words masked out could successfully distinguish 64,248 Amazon users with an F1 score of about 0.79 in a verification setup. 
\citet{narayanan2012feasibility} reported a performance of over 20\% accuracy when ascribing texts among 100,000 candidates. 
Studies from large corpora suggest that writing styles are distinguishable at scale.

% evidence supporting one's style is somehow consistent
Consistency in an author's style refers to the recurring appearance of certain linguistic patterns with a relatively stable frequency in their writing.
However, the consistency of writing style shown in one's documents is not absolute.
From a statistical perspective, the stylistic measures of a new document do not deviate significantly from those of the writer's previous works, as long as there are no significant differences in the underlying influential factors.
For machine learning, the ideal situation is that the training and testing samples come from the same distribution or share many overlapping attributes, such as topic and genre.
\citet{overdorf2016blogs} observed that an SVM model performs better when the ``gap'' between the samples in the training set and those in the testing set, as measured by feature vectors, is smaller.
In practice, it is hard to imagine that a stylistic classifier predicated on perfect alignment between training and testing data can be of wide use.

\section{Adversarial Stylometry\label{sec: adversarial_stylometry}}

\subsection{A Brief History\label{sec: as_history}}
% a bit of legal and techniqucal backgrounds of anonnymity and its protection
The right to speak anonymously is deeply rooted in many democratic societies.
The European Convention on Human Rights staunchly defends personal privacy, with its eighth article often cited in litigation cases involving mass surveillance.
The General Data Protection Regulation (GDPR), enacted in 2018, grants Europeans the ability to access, delete, and correct personal information that companies store about them, thus elevating anonymity and pseudonymity to new heights.
The First Amendment to the United States Constitution protects free speech, and courts in the United States have also recognized a protection for anonymous speech, although there are limits to this protection.
For example, if someone is engaging in illegal activity or defamation, a court may order that their identity be revealed.
% The First Amendment to the United States Constitution protects free speech by preventing the government from unmasking individuals.
% This tradition may trace back to colonial times, when the country greatly benefited from dissenters who expressed their political opinions through pamphlets.

Even authoritarian regimes, to some extent, respect free speech, but they are not friendly towards anonymous speech. 
China recognizes free speech in its Constitution, though it imposes restrictions through its Cybersecurity Law and other regulations, such as requiring social media users and domestic website hosts to register under their real names, among other efforts to enforce identity disclosure.
Similarly, Russian government mandates foreign instant messaging services to verify a user's identity through their phone number within 20 minutes of a request, disallowing unidentified users from registering or remaining on the platform.
Surveillance capitalism poses its own risks to online anonymity despite the existence of protective laws and technologies.
The First Amendment to the US Constitution regulates state actors but leaves space for private sector activity, allowing surveillance capitalism to infiltrate.
Tech giants, holding vast amounts of user data, can perform actions covertly, including breaches of identity.

Technologies for online anonymity have flourished in the past several decades, ranging from general tools such as VPNs and the Onion Router to apps supporting end-to-end encryption. 
However, people remain largely unaware of the risks posed by their individual stylistic nuances, which could potentially render these channel protection efforts useless.
In what follows, we provide a historical overview of stylometric analysis, tracing its roots from humanities to privacy concerns, and discuss the development of countermeasures to address these concerns.

\subsubsection{Evolution of Stylometric Analysis}
% a bit of history of stylometry
% the area stems from humanities interests
Stylometry originated from an interest in the humanities, with early works primarily focused on resolving authorship disputes in historical documents.
Notable examples include studies on the \emph{Federalist Papers} \citep{mosteller1964inference}, \emph{Letters of Junius} \citep{ellegrd1962julniusletter}, and disputed works of Shakespeare \citep{mendenhall1901mechanical, williams1975mendenhalls}, among others.
While there are discussions about the use of hand-crafted rules to mimic the style of other novelists \citep[Ch. 6]{hoover1999language}, and the stylistic resemblance between original works and their pastiche versions \citep{somers2003authorship}, the scope of stylometry has long remained within the realm of historical and literary studies.

% forensics and fingerprinting famous authors
Stylometry gained considerable attention as a forensic tool in the 1990s, highlighting its potential applications beyond the scope of humanities.
The most renowned case in this regard is that of the Unabomber, a domestic terrorist who targeted individuals involved in modern technology.
The Unabomber was eventually apprehended when his brother informally observed that the Manifesto bore his writing style and proposed an investigation.
Stylometric analyses began to be admitted as evidence in courts in the US and UK.
While less reliable measures and reported failed forensics \citep[see][Sec. 2.3.3 for discussion]{juola2008authorship} have since largely hindered the widespread legal admissibility of stylometric analyses, rigorous analyses have received testimonials over time.
For instance, in 2013, the detective novel \emph{The Cuckoo's Calling}, published under the pseudonym Robert Galbraith, was suspected to be the work of J. K. Rowling, author of the \emph{Harry Potter} series.
The authorship was determined by comparing the most common words and character four-grams of the book with Rowling's other works. 
Rowling later confirmed that she was indeed the author.

% fingerprinting random individuals
In the 2000 USENIX Security Symposium, \citet{rao2000can} first expressed concerns that when applied inappropriately, stylometry could infringe on the anonymity of internet users.
After removing explicit identity-related information, the authors identified 68 users who had posted more than 50 posts, accumulating at least 5,000 words in total. 
They split each user's writing in half and projected the frequency of generic words onto a coordinate space using principal components.
They discovered that out of the 68 pairs from the same user, 40 were closest to each other.
At this point, it became clear that personal stylistic nuances could be leveraged to undermine online anonymity.
This threat to blog post anonymity persists even when using writing samples from a distinct domain, i.e., emails.
% more powerful algorithm and real world applications
Later, empirical studies confirmed the feasibility of large-scale fingerprinting, ranging from ten thousand \citep{koppel2011inthewild} to a hundred thousand bloggers \citep{narayanan2012feasibility}. 
Similar approaches have been used to successfully identify sock puppet accounts in underground forums with high performance \citep{afroz2014doppelganger}.

\subsubsection{Rise of Adversarial Stylometry}
Given these developments, it was a matter of emergency to create a defense to thwart stylometric analysis.
The research on countermeasures began around 2006.
\citet{kacmarcik2006obfuscating} investigated whether it would be possible to circumvent standard authorship attribution techniques.
Using the Federalist Papers as a test bed, the authors demonstrated that manipulating fourteen words per thousand is sufficient to alter the conclusions of a standard authorship attribution method, which would otherwise accurately predict the authorship.
However, because they directly manipulated the vector space of common word frequencies, the effectiveness of their approach remains unclear when operationalized in the symbolic space.

\citet{brennan2012adversarial} introduced the term ``adversarial stylometry'' to describe the defense as ``applying deception to writing style to influence the outcome of stylometric analysis.''
Other terms have been used to refer to these and related techniques, including author obfuscation/masking \citep{potthast2016author}, deceptive style \citep{juola2012detecting}, anonymous authoring \citep{le2015secure}, author(ship) anonymization \citep{almishari2014fighting, bo2020authorship}, author concealment \citep{day2016towards}, and authorship privacy technologies \citep{hiatus}.

Adversarial stylometry has received attention from researchers both within and beyond stylometry.
A significant event took place between 2016 and 2018 when the Uncovering Plagiarism, Authorship, and Social Software Misuse (PAN) lab at the Conference and Labs of the Evaluation Forum (CLEF) proposed shared adversarial stylometry tasks.
These tasks were designed to challenge the state-of-the-art models proposed to solve the previous years' shared authorship verification problems. 

\subsubsection{Joint Advances with Differential Privacy\label{sec: dp}}
% the join of privacy community with interests in generating privacy-preserving text, local DP in particular
A closely related research line contributing to the defense against authorship identification attacks originates from the PPDP community, and from differential privacy in particular.
In a typical setup, PPDP works with personal data contained in a \emph{table}.
These approaches aim to minimize the risk of personal privacy violations by adding noise and generalizing personally identifiable information (PII), subsequently releasing only noise-altered and irreversible data to the public.
Although personal written documents are not explicitly mentioned in the PII examples provided by the National Institute of Standards and Technology (NIST), which include items such as names, fingerprints, and passport numbers, they still qualify as PII according to NIST's definition of PII as ``information about an individual that is linked or linkable to one of the above'' \citep{mccallister2010guide}.
The original framework of differential privacy, designed for sanitizing tabular data, presents challenges when it comes to providing privacy guarantees for a single personal written document, which is our primary focus. 
Below, we outline the typical setting of differential privacy before relating its principles to the defense against authorship identification attacks.

\paragraph{A Primer of Differential Privacy}
The rationale of differential privacy can best be illustrated with an example.
Consider a dataset $x_1 \in \mathcal{X}$ that contains records of a group of patients' infection statuses in bits and a query on the dataset $f: \mathcal{X} \rightarrow \mathbb{R}$ that returns the count of positive cases.
An adversary could easily determine the status of patient $A$ by comparing the outputs of $f(x_1)$ and $f(x_2)$, where $x_2$ retrieves the summary of all records from $x_1$ except for $A$.

This concern can be mitigated with a defensive \emph{randomized algorithm} $\mathcal{M}: \mathcal{X} \rightarrow \mathcal{Z}$, where the input is perturbed with noise drawn from a certain probability distribution.
In the strictest sense, any pairs of datasets differing by only one entry should yield similar summaries that are indistinguishable after the transformation.
These datasets are deemed \emph{neighboring} because they differ only in one record.
We say a randomized algorithm $\mathcal{M}$ satisfies $(\epsilon, 0)$-differential privacy if and only if for any neighboring datasets $x_1, x_2 \in \mathcal{X}$, the following holds:
\begin{align}
Pr[\mathcal{M}(x_1)=z] \leq exp(\epsilon) \cdot Pr[\mathcal{M}(x_2)=z]  
\end{align}

This means that to protect each individual’s privacy, differential privacy injects noise into the output such that the results from any neighboring dataset sufficiently resemble each other, as their difference is bounded by $\exp(\epsilon)$ \citep{dwork2014algorithmic}.
The defense mechanism is parameterized with the \emph{privacy loss budget} $\epsilon$ representing the maximum allowable privacy loss or the extent to which an individual's data can be compromised.
This budget governs the trade-off between privacy and data utility, where a smaller $\epsilon$ indicates stronger privacy protection but potentially reduced data utility, and vice versa.

A typical differential privacy strategy unfolds as follows.
First, the maximum difference of a query between any neighboring datasets, known as \emph{sensitivity}, is calculated.
By quantifying sensitivity, we can determine an upper bound on the amount of noise required.
With a pre-defined privacy budget, the appropriate noise mechanism is then selected.
For instance, the Laplace or Gaussian mechanism samples noise from the Laplace or Gaussian distribution with a scale set by the sensitivity and privacy budget.
The exponential mechanism is used for more complex queries, where noise is generated based on the likelihood of the query's output.

A comprehensive review of differential privacy is beyond the scope of this work. 
However, for those interested in differential privacy, we recommend the NLP researcher-friendly introduction by \citet{igamberdiev2021privacy}, as well as the tutorial from \citet{vadhan2017complexity} for detailed explanations of its mechanisms and proofs.

\paragraph{Relating to Adversarial stylometry}
The potential to adapt the canonical differential privacy setup to our goal of preserving document anonymity exists, but it requires two major adjustments.

First, in a typical setup, the records are collected by a trusted central node, known as a \emph{curator}.
In our case, we can assume that the table contains only one entry, the document whose anonymity an adversary wants to compromise.
Since neighboring datasets require a one-entry difference, we can treat any single piece of personal writing as neighboring to any other. 
This setup removes the need for a curator and is referred to as \emph{local differential privacy}. 
Local differential privacy is relevant to situations where an adversary has extensive knowledge of potential authors and the potential adversarial stylometry methods they may employ to evade fingerprinting.

Second, applying differential privacy mechanisms to text data is not as straightforward as it is with numeric and Boolean data.
Since 2018, two lines of research have emerged to address the challenge of anonymizing personal documents under the framework of local differential privacy.
The first approach operates at the token level.
Each token is perturbed using its numerical representations, often derived from a static or contextual word embedding.
These perturbed tokens are then projected to other related tokens based on a certain distance measure.
The cumulative privacy loss can be calculated with the post-processing property to fulfill a privacy budget. 
We will address this in Section~\ref{sec: syonoym_substitution}.
The second approach typically starts by encoding the entire text into a fixed-length numerical representation. 
Subsequently, calibrated noise is injected into the representation or realized via generation sampling. 
Works along this line will be discussed in Section~\ref{sec: style_transfer}.

\subsection{Formalization\label{sec: formalization}}
Adversarial stylometry is a defense mechanism that seeks to obscure the style of a document while ensuring successful communication in natural language.
To ensure clarity in terminology, we refer to entities conducting authorship identification tasks as \emph{adversaries}, and the corresponding model is referred to as a \emph{threat model}.

Recall that a threat model is trained to find a parameter set that minimizes the loss between the true authors and the predicted authors (cf. Eq.~\ref{eq: stylometry_as_classification}).
During the testing phase, for a given document $\bm{x}_k$ whose authorship is of interest, $f_{\bm{\theta}}(\bm{x}_k)$ represents the model's decision, be it the predicted identity in an attribution task, or whether the samples were written by the same author in verification.
Here, $\bm{x}_k$ is from the test set $D_s$ and corresponds to the true label $y_k$.
In order to confound a threat model, adversarial stylometry approaches manipulate the test sample to influence the prediction.
The test sample $\bm{x}_k$ is perturbed by a noise $\delta \bm{x}$ to create an adversarial sample $\bm{x}_{k}^{\prime}$ such that $f_{\bm{\theta}}(\bm{x}_{k}^{\prime})$ does not match $y_{k}$.
In formula,
\begin{equation}\label{eq: adversarial_stylometry}
f_{\bm{\theta}}(\bm{x}_k^{\prime}) = y_{k}^{\prime}, \textrm{where}
\ \bm{x}_{k}^{\prime} = \bm{x}_{k} + \delta \bm{x}, \ y_{k}^{\prime} \neq y_{k}
\end{equation}
In attribution problems, if $y_{k}^{\prime}$ represents a specific author that the model is manipulated to favor, it is considered a \emph{targeted} approach (or ``framing'').

\subsection{Taxonomy\label{sec: taxonomy}}
The goal of adversarial stylometry is to disguise an author's authentic writing $\bm{x}_k$ with an adversarial sample $\bm{x}_{k}^{\prime}$ by a perturbation $\delta \bm{x}$.
Considering the source, direction, and magnitude of perturbations, we highlight three axes in our taxonomy.

\paragraph{Automation}
Adversarial stylometry approaches are traditionally classified based on the degree of automation involved in obtaining the perturbation: \emph{manual}, \emph{computer-assisted}, and \emph{automated} approaches \citep{potthast2016author}.
This categorization roughly corresponds to the source of intelligence inputs.
Within the category of fully automatic evasion, we further distinguish \emph{modification-based} and \emph{generation-based} approaches to reflect distinct methodological frameworks: the modification-based methods operate in the discrete, symbolic space, while generation-based approaches work with numerical representations before discretization.

\paragraph{Favored Style}
Existing approaches can be characterized as targeted or untargeted based on whether they favor a particular style (see Eq.~\ref{eq: adversarial_stylometry}).
While untargeted approaches modify the authorial style away from its canonical form, targeted approaches favor a specific style, which can be from an author, from a sub-population within an entire corpus, or defined in a data-driven manner.

\paragraph{Availability of Background Knowledge}\label{sec: taxonomy_available_knowledge}
Alternatively, a taxonomy can be developed based on the availability of background knowledge of the document from the adversary's point of view.
Intuitively, more such information about a case (e.g., the model architectures, parameters, gradients, or corpora in use) necessitates the injection of a larger amount of noise.
Differential privacy derives the notion of a privacy budget in an information theoretic manner; hence, it corresponds to a situation in which everything but the author's identity is known to an adversary.

\subsection{Considerations \& Evaluation\label{sec: evaluation}}
The most crucial factor in evaluating the quality of an adversarial stylometry approach is its \emph{effectiveness} in inducing a performance drop in the relevant threat models.
However, while simply scrambling the document with random words may evade authorship fingerprinting, it will also fail to convey the intended message.
Instead, an ideal adversarial sample should not only evade authorship fingerprinting but also facilitate successful communication.

\subsubsection{Effectiveness}
The metric of \emph{effectiveness} gauges the adversarial capacity of a defense to evade fingerprinting.
It is typically measured by the change in performance of an adversary caused by an adversarial sample: an effective defense model is expected to decrease the performance of a threat model.
For corpora where each class has comparable amounts of samples, the performance of an adversary is often reported using accuracy.
For unbalanced corpora, it is common to report an array of measures including precision, recall, F-scores, and ROC-AUC curves;
the commonly used Matthews correlation coefficient (MCC) \citep{matthews1975mcc, gorodkin2004mcc} takes into account true positives, true negatives, false positives, and false negatives to calculate a value between $-1$ and 1, where 1 indicates perfect prediction, 0 indicates random prediction, and $-1$ indicates total disagreement between the model and the actual data.

A promising measure for comparing defense effectiveness, termed the \emph{world ranking score}, is proposed by \citet{potthast2018overview}.
This score employs 44 threat models previously submitted to the PAN shared tasks of authorship verification as the adversary.
It takes into account the effectiveness and inductive bias of threat models as well as the difficulty of samples.
For researchers who prefer a decentralized measure, easy-to-reproduce threat models, such as a linear SVM based on function words, can serve as a viable substitute.

\paragraph{Stylistic Divergence}
Instead of reporting the impact on a threat model, it is intuitive to measure the stylistic distance derived from the divergence between the perturbed sentence and its source.
A common measure is PINC (Paraphrase In N-gram Changes) \citep{chen2011collecting}, which quantifies lexical dissimilarity by rewarding the introduction of new n-grams that appear in the perturbed sentence but not in the source.
% two similar yet older, worse metrics: 
% Parametric: Anautomatic evaluation metric for paraphrasing
% PEM: A paraphrase evaluation metric exploiting parallel texts

\emph{Ad hoc} measures have also been proposed to address stylistic distance.
Normalized cosine distance has been used to gauge stylistic distance in the space of word n-grams \citep{xu2012paraphrasing}, as well as the Writeprints variant \citep{overdorf2016blogs}.
Variants of Jensen-Shannon divergence have also been reported.
\citet{backes2016zoos} reported the square root of Jensen-Shannon divergence over word uni-grams as a stylistic distance measure.
Similarly, \citet{bevendorff2019heuristic} measured stylistic distance using Jensen-Shannon divergence over the character tri-gram space and found that the square root of Jensen-Shannon divergence is inversely correlated with 2-based logarithmic document length measured in characters.

\subsubsection{Semantics Retention}
A perturbed sample should maintain semantic coherence with the original document.
The similarity between the original and perturbed samples is often measured either on the surface level or in a vector space.
Representative measures of surface value include the Bilingual Evaluation Understudy (BLEU) and the Metric for Evaluation of Translation with Explicit ORdering (METEOR).
BLEU counts the number of overlapping n-grams between the perturbed sample and the original document, then normalizes this count by the total number of n-grams.
A major limitation of BLEU is that it does not consider the semantic equivalence of phrases or sentences.
Thus, BLEU can penalize valid, semantically equivalent translations that do not use exactly the same words or phrases as the original.
METEOR considers precision, recall, and synonymy, thereby providing a more comprehensive evaluation of semantic similarity.
It also accounts for differences in word order and uses stemmed versions of words and paraphrases, which increases its flexibility.

Semantic distances can also be computed with static and contextual embeddings, such as the Word Mover's Distance (WMD) \citep{wmdkusnerb15}, BERTScore \citep{zhangbertscore}, and the Universal Sentence Encoder (USE) \citep{cer2018universal}.
WMD utilizes word embeddings such as word2vec and calculates the minimum cost to relocate words from one document to match the position of words in the other, thereby quantifying their semantic similarity.
BERTScore and USE represent documents as contextual embeddings and compute the cosine similarity between these embeddings for a pair of texts.
The newer measures arguably capture nuanced linguistic similarities better than traditional n-gram-based metrics.
However, because they incorporate components based on neural networks, they present challenges in reproducibility and comparison.
% TODO: Decide which category \citet{wu1994verb} falls in, it is a ``semantic similarity metric'' by \citet{grondahl2019text}
% TODO: Latent semantic analysis
% TODO: QA as a way to check informativeness

\subsubsection{Grammaticality}
Perturbed samples ought as well to maintain fluency and grammaticality to obviate subsequent investigation.
Perplexity is a commonly used metric for grammaticality, based on the heuristic that lower perplexity implies a better fit of the model's probability distribution to the training corpora.
Essentially, it quantifies the uncertainty of the language model in predicting the subsequent word in a sequence.
However, while perplexity is a useful proxy, it does not directly measure grammaticality.
Factors such as the similarity between the training and test data in terms of domain and style, as well as the degree to which the language model is well-trained, significantly impact its value.
For instance, a fluent perturbed sample with an informal tone is likely to receive a high score when evaluated with a language model trained on formal corpora.

A grammatical acceptability classifier can also be utilized. 
These classifiers can be trained on the Corpus of Linguistic Acceptability (CoLA) \citep{warstadt2018neural}, a dataset comprising sentences labeled for their grammatical acceptability.
CoLA includes both grammatical and ungrammatical English sentences, sourced from linguistic literature; it offers a more direct measure of grammaticality compared to perplexity.
Nevertheless, performance on CoLA depends on the quality and representativeness of the training data, and models may not generalize well to sentence types not included in the dataset, e.g., those with an informal tone.

\subsection{Other Considerations\label{sec: other_considerations}}
Apart from the three frequently reported aspects above, we further elaborate on four additional attributes that a good defense should possess: \emph{inconspicuousness}, \emph{irreversibility}, \emph{transferability}, and \emph{explainability}. 
Though these criteria are desirable, failure to meet them often does not pose an immediate threat to anonymity.

\subsubsection{Inconspicuousness}
Inconspicuousness requires that the defense operate subtly and leave little trace of its intervention.
As with grammaticality, it is preferable not to alert readers or stylometric analysis tools to any alterations.
This is particularly important for fully automatic approaches.
% todo: juola2011analyzing, afroz2012detecting found manual obfuscation is easy to detect
Back-translation is a straightforward method of disrupting one's style, yet statistical machine translation has been reported to leave discernible fingerprints \citep{caliskan2012translate}.

\subsubsection{Irreversibility\label{sec: irreversibility}}

If the transformation can be reversed, even partially, the probability of exposing one's writing style increases.
Well-crafted transformations should be irreversible to ensure that original authorship patterns cannot be restored by a human reader or an automatic de-obfuscator.
Irreversibility is associated with three factors: the complexity and magnitude of the transformation, as well as prior knowledge of a defense.

% complexity and scale
The more complex a perturbation, the less likely it is to be reversed.
In this regard, transformations from a deterministic or a simple stochastic distribution are particularly susceptible to reversal.
Therefore, rule-based evasion should be considered less safe in the face of reversing efforts.
A common misconception in crafting adversarial examples, as pointed out by \citet{biggio2018wild}, is that adversarial examples should be minimally perturbed.
Although minimizing perturbation is often a strategy to reduce computation cost and maintain readability during the modification of the original sample, minimizing perturbation magnitude \emph{per se} is tangential to the objective of effectiveness.
Perturbation on a higher scale, possibly resulting from a diction- and sentence-level reorganization as well as lexical-level disruption, is potentially a more powerful defense.

% a-prior knowledge
It is intuitive that a perturbed sample is vulnerable to a threat that has partial or even perfect knowledge of the obfuscating method.
Assuming the training corpus and heuristic are available, \citet{le2015secure} have demonstrated that defenses employing minimal randomness \citep{kacmarcik2006obfuscating, mcdonald2012use} could be reversed. 

\subsubsection{Transferability}
Transferability refers to the ability of a defense, designed to mislead certain stylometric classifiers, to successfully deceive other classifiers.
This concept is crucial as it suggests that the defense, often without access to the knowledge of an adversary, could potentially evade authorship identification by creating adversarial examples based on a different, accessible model.

\citet{haroon2021avengers} proposed an ensemble-based method to enhance the transferability of a heuristic-based approach that relies on an internal stylometric classifier to select appropriate synonyms \citep{mahmood2019mutantx}.
Transferability was measured by the percentage of obfuscated documents produced by an internal classifier that were misclassified by an adversary classifier.
Their experiments demonstrate that incorporating the decision boundaries of multiple underlying attribution classifiers is more likely to be effective against different attribution classifiers.
Importantly, this improved performance persists even when the adversary's classifier operates in a different feature space. Similarly, \citet{zhai2022girl}, \citet{mattern2022limits}, and \citet{grondahl2020effective} have found that threat models trained in an adversarial manner can significantly diminish the effectiveness of existing obfuscation techniques.

\subsubsection{Explainability\label{sec: explainability}}

Explainability is essential for establishing trust and ensuring the reliable and ethical use of adversarial stylometry techniques.
Making defenses interpretable enables users to understand why and how a perturbed sample is obtained.
Further knowledge of which words are being adjusted to conceal writing style can aid users in maintaining their anonymity more effectively.

Explainability also ensures adversarial stylometry does not exhibit bias towards specific authors or writing styles.
This precaution protects against potential misuse or skewed results that might disadvantage certain individuals or groups.

Despite its significance, the aspect of explainability remains relatively underexplored in adversarial stylometry.
In contrast, more established NLP domains frequently utilize \emph{post hoc} methods such as Local Interpretable Model-Agnostic Explanations (LIME) \citep{ribeiro2016should}, SHAP (SHapley Additive exPlanations) \citep{lundberg2017unified}, Anchors \citep{ribeiro2018anchors}, and counterfactual frameworks \citep{ge2021counterfactual}. 
For research employing attention-based models, attention weights can serve to explain the parts of the input on which the model focused to derive its predictions \citep{clark2019does}.
More recently, architectures that maintain full mathematical interpretability and effectively compete with well-engineered Transformers have become available \citep{yu2023white}.

\subsection{Human Evaluation}

While automated techniques can swiftly process vast amounts of data, they might overlook subtle nuances of style that a trained human eye can discern.
Human evaluation plays a vital role in adversarial stylometry by ensuring the effectiveness and quality of interventions.
However, the challenges of involving human evaluators are well-known: it is labor-intensive and costly, and forensic linguists might hesitate due to potential reputation risks.
Despite these challenges, researchers are advised to conduct a reasonably scaled user study to assess semantic relevance, grammaticality, and inconspicuousness.
Reporting representative successful and failed perturbed samples can also provide a source of insight.

\section{Manual Evasion\label{sec: manual}}

Manual interventions posit that a motivated individual is capable of sufficiently altering their prose style in a new document such that stylometric adversaries struggle to link the document to the individual's previous writing.
The primary methods are \textit{obfuscation} and \textit{imitation}. 
Though limited in number, these approaches have shown high effectiveness in challenging various authorship attribution techniques.
A summary of the studies surveyed in this section is presented in Table~\ref{tbl: manual_summary}.

\begin{table}[!ht]
 \caption{Summary of manual evasion methods. Semantic relevance and grammaticality are not listed as the samples are composed by humans. BG, EBG, and RJ stand for the Brennan-Greenstadt, Extended Brennan-Greenstadt, and Riddell-Juola corpora, respectively.
 \label{tbl: manual_summary}}
 \small
  \centering
    \resizebox{\textwidth}{!}
  {
  \begin{tabular}{p{3cm}p{1.5cm}p{5cm}p{1.8cm}p{3cm}p{5cm}}
    \toprule
    Work
    & Adversary
    & Threat Models
    & Defense
    & Corpora
    & Effectiveness \\ \midrule 
    \citet{brennan2009practical}
    & Attribution
    & Chi-squared (word lengths, character uni-gram, and punctuation), MLP (Basic-9), \& synonym-based
    & Obfuscation \& imitation
    & BG 
    & Near-chance accuracy for obfuscation; below-chance for imitation \\ \cmidrule(r){1-6}
    \citet{juola2010empirical} 
    & Attribution
    & Euclidean histogram distance, Manhattan distance, cosine distance, \& linear discriminant analysis
    & Obfuscation \& imitation
    & BG
    & Above-chance accuracy by a significant margin in most cases for obfuscation; mostly below-chance for imitation \\ \cmidrule(r){1-6}
    \citet{brennan2012adversarial} 
    & Attribution
    & SVM (Writeprints-static), MLP (Basic-9), \& synonym-based heuristics
    & Obfuscation \& imitation 
    & EBG
    & Mostly below-chance \\ \cmidrule(r){1-6}
    \citet{almishari2014fighting} 
    & Attribution
    & Chi-squared (character tri-gram and POS bi-gram)
    & Crowdsourcing rewrite
    & Yelp 
    & Substantive accuracy drop, yet above-chance by a large margin \\ \cmidrule(r){1-6}
    \citet{wang2022reproduction} 
    & Attribution
    & SVM (Writeprints-static), RoBERTa, \& logistic regression (function word) 
    & Obfuscation \& imitation
    & EBG \& RJ
    & Substantive accuracy drop; above-chance by a discernible margin in most cases\\
    \bottomrule
  \end{tabular}
  }
\end{table}

\subsection{Obfuscation}
Obfuscation refers to situations in which an author intentionally writes differently than they ordinarily would.
Several field studies have reported on the effectiveness of obfuscation.
\citet{brennan2009practical} recruited twelve laypeople, each providing approximately 5,000 words of their previous formal writing as training data.
Participants then wrote a 500-word description of their neighborhood for a friend unfamiliar with it, attempting to ``hide their identity through their writing style.''
The resulting training and test samples form the Brennan-Greenstadt corpus.
Three threat models were implemented: a Chi-squared test based on word lengths, character uni-grams, and punctuation; a synonym-based model; and a multilayer perceptron utilizing complexity measures.
In a cross-validation setting on the training set, the accuracy across threat models rarely fell below 80\% when considering 2 to 5 random candidates.
However, the accuracy with newly-written responses was at the chance level across threat models.

\citet{juola2010empirical} revisited the Brennan-Greenstadt corpus, examining 160 combinations of preprocessing (such as enforcing lowercase, removing punctuation, and normalizing multiple whitespace characters to one), features (e.g., function words and character bi-/tri-grams), and algorithm choice (i.e., Euclidean histogram distance, Manhattan distance, cosine distance, and linear discriminant analysis).
The accuracy of the threat models was found to be higher than chance by a noticeable margin in general.
A potential explanation for this discrepancy is that character tri-grams, the most informative feature in \citet{juola2010empirical}'s study, were not used by \citet{brennan2009practical}.

In a replication study with 45 participants recruited from Amazon Mechanical Turk (MTurk), \citet{brennan2012adversarial} requested 6,500 words of pre-existing samples and followed the same procedure to solicit writing responses, forming the Extended Brennan-Greenstadt corpus.
They replaced the Chi-squared method with a polynomial SVM based on a variant of the Writeprints feature set \citep{abbasi2008writeprints}.
This feature set, termed Writeprints-static, includes a range of feature categories, such as character n-grams, function words, POS n-grams, and complexity measures.
The results were consistent with the previous study: with candidate sets ranging from 5 to 40, only the SVM performed marginally better than chance.
In both experiments, participants appeared to ``dumb down'' their writing by using less descriptive words and shorter sentences.

\citet{almishari2014fighting} investigated the feasibility of using MTurk workers to rewrite Yelp user reviews.
When the test samples were rewritten by MTurkers, the accuracy of the Chi-squared test (using character tri-grams and POS bi-grams as features) dropped from 95\% to 55\%.
Semantic retention and grammaticality, evaluated using a different population of MTurk workers, were deemed satisfactory.
Manual examination of the rewrites revealed that new styles were often introduced without substantial changes to the content structure, primarily at the sentence level.
While this approach requires extra security precautions when communicating with MTurk workers, it provides evidence that sentence-level transformations can sufficiently influence the decisions of an attribution threat.
The risk of identity disclosure in the obfuscated documents was significantly higher (e.g., an accuracy of 55\% when considering 40 candidates and 10\% when considering 1,000 candidates), compared to the chance-level performance reported in \citet{brennan2009practical} and \citet{brennan2012adversarial}.
The main reason is arguably the significantly larger amount of training data used in \citet{almishari2014fighting} (on average, 14,751 words per user, twice as much as in \citet{brennan2012adversarial}).
The differences in preprocessing, models' inductive biases, and features relied upon may also contribute to the observed performance discrepancies.

In a recent study reproducing the results with the Extended Brennan-Greenstadt corpus, \citet{wang2022reproduction} found that, with proper normalization and standardization of the Writeprints-static feature space, an SVM threat model could still perform significantly better than chance level (e.g., approximately 35\% accuracy when considering 5 candidates and about 15\% among 40 authors).
The authors reported performance from three threat models: a RoBERTa, a polynomial SVM, and a logistic regression using 512 function words as features.
They replicated the experiment with a new population of MTurk workers, with each participant randomly assigned one writing prompt, creating the Riddell-Juola corpus.
They found the obfuscation method effectively penalized all three threat models to a sufficient degree.
Contrary to \citet{brennan2012adversarial}'s observation, they also reported that the simple logistic regression's performance was above chance by a discernible margin in both the reproduction and replication studies.

\subsection{Imitation}
Imitation involves an author adopting the style of another writer.
This approach was also introduced by \citet{brennan2009practical}.
Utilizing the same population that performed the obfuscation task, the authors tasked the MTurk workers with reading a 2,500-word excerpt from \textit{The Road} by Cormac McCarthy.
The reading material was selected due to McCarthy's minimalist and starkly poetic writing style.
Subsequently, participants were asked to describe their day from the time they got out of bed, using a third-person perspective.
Both \citet{brennan2009practical} and \citet{juola2010empirical} reported threat models' accuracy to be below chance level when using the Brennan-Greenstadt corpus, as well as in a subsequent replication using the Extended Brennan-Greenstadt corpus.
The imitation of Cormac McCarthy's style was successful; half the time, the multilayer perceptron and synonym-based models incorrectly attributed the imitation prose from other authors to Cormac McCarthy.
This indicates that a layperson can alter their style significantly enough to resemble a distinct style after relatively brief exposure to it.
\citet{brennan2009practical} reported the use of descriptive and grim language in the imitation samples.
However, a recent replication by \citet{wang2022reproduction} suggested that this result may have been somewhat exaggerated: a logistic regression threat model employing function words and a RoBERTa-base performed better than chance by a margin.
\citet{wang2022reproduction} reported that the obfuscation method was slightly more effective than the imitation method in the Riddell-Juola corpus.

\subsection{Summary}
% pro: effective; low-resource; guaranteed grammaticality and fluency
Overall, the effectiveness of manual approaches is recognized.
Since these are composed by human subjects, they guarantee semantic relevance and appropriate language use.
Moreover, manual interventions to obscure one's style merit attention, even if they prove less effective than machine-assisted style obfuscation methods, as they can be used when trusted computational resources are unavailable.
Such a scenario is conceivable in the case of whistle-blowing.
For instance, an employee at a financial firm looking to expose malfeasance while maintaining anonymity may lack immediate access to an unmonitored computer.

% con: not diverse; debatable method effectiveness; long-term protection
Despite its merits, manual intervention is not as diverse as defenses involving algorithms; it comprises only two approaches and has not been tested against verification threats.
Moreover, the relative effectiveness of obfuscation and imitation is inconsistently observed across the (Extended) Brennan-Greenstadt and Riddell-Juola corpora, necessitating further investigation.
Due to the labor-intensive nature of data collection, conducting such field studies can be prohibitively expensive.
Lastly, as writing style is habitual and can be considered a behavioral biometric, it is reasonable to argue that long-term anonymity could be challenging to maintain through manual modifications of style alone \citep{afroz2012detecting}.

\section{Human-Machine Collaborative Evasion}\label{sec: human_computer_collaboration}
Human-machine collaboration requires the participation of both human beings and models, with various degrees of human involvement.
Often, a software application is developed to assist in the rewriting process by offering users tailored feedback on modifications in phrasing, or by prompting a user to choose from a collection of rewrites, procedures which naturally raise considerations of end-user delivery, user interface, and experience design.
This research direction is closely aligned with the ``computer-assisted obfuscation'' approaches described in \citet{potthast2016author}; we rebrand it as human-machine collaboration to encompass the variety of roles humans play and the research opportunities available.
A summary of the studies surveyed in this section is presented in Table~\ref{tbl: human_machine_collaboration_summary}.

\begin{table}[!ht]
 \caption{Summary of human-machine collaborative evasion methods. ``Constraints'' notes whether a defense considers factors other than effectiveness; ``Targeted'' denotes whether a defense is specifically tailored to a particular style.
 \label{tbl: human_machine_collaboration_summary}}
 \small
  \centering
      \resizebox{\textwidth}{!}
  {
  \begin{tabular}{p{3cm}p{1.5cm}p{3.1cm}p{1.8cm}p{2cm}p{4cm}p{1.5cm}p{1.5cm}}
    \toprule
    Work
    & Adversary
    & Threat Models
    & Defense
    & Corpora
    & Effectiveness
    & Constraints
    & Targeted \\
    \midrule
    Anonymouth \citep{mcdonald2012use}
    & Attribution
    & SVM (Basic-9)
    & Rewrite by oneself
    & Anonymouth-10
    & 70\% accuracy drop (10 authors)
    & Semantic \& grammatical
    & No \\ \cmidrule(r){1-8}
    \citet{le2015secure}
    & Attribution
    & SVM (Basic-9)
    & Rewrite by others
    & Anonymouth-10
    & Below random chance
    & Semantic \& grammatical 
    & No \\ \cmidrule(r){1-8}
    AuthorCAAT \citep{day2016towards}
    & Attribution
    & Manhattan distance (character uni-grams)
    & Back-translation
    & CASIS
    & 48\% accuracy drop (100 authors)
    & Semantic 
    & No \\ \cmidrule(r){1-8}
    AuthorCAAT-III \citep{faust2017adversarial}
    & attribution
    & SVM/MLP (Writeprints Limited/Basic-9)
    & Back-translation
    & CASIS
    & 66.7\% accuracy drop on SVM (Writeprints-limited, 25 authors)
    & Semantic 
    & Yes \\ \cmidrule(r){1-8}
    AIM-IT \citep{allred2020towards}
    & Attribution
    & Cosine distance (character 4-grams), normalized character bi-grams, suffix-based measure, \& character CNN 
    & Back-translation
    & CASIS
    & 12\% accuracy drop on character CNN (25 authors )
    & Semantic 
    & Yes \\
    \bottomrule
  \end{tabular}
  }
\end{table}

Anonymouth is a tool designed to obfuscate the style of written text, aligning it with a general population by proposing user modifications to minimize identification risk \citep{mcdonald2012use}.
Anonymouth operates by first identifying the most discriminating features that set the text to be anonymized apart from the background population.
Each of these features is then clustered using k-means.
User feedback is generated based on a ``generic'' cluster's centroid, with evaluations based on the number of samples in the cluster and its distance from the document to be obfuscated.
\citet{mcdonald2012use} noted that users had difficulty following the suggestions when applying inclusive feature sets, such as a variant of Writeprints.
Therefore, the authors opted for a limited feature set containing only nine complexity measures, the ``Basic-9'' feature set.
Despite the fact that the edited texts displayed strong adversarial characteristics by reducing the accuracy of a Basic-9 featured SVM in a field study, these texts have not yet been tested against threat models relying on more comprehensive feature sets.
It was also observed that users complicated their prose when following Anonymouth's feedback, as indicated by increased sentence length and lower readability.
This contrasts with observations on manual obfuscation where subjects tended to ``dumb down'' their writings \citep{brennan2009practical, brennan2012adversarial}.

A newer version of Anonymouth is able to suggest paraphrased alternatives by first translating a sentence into an intermediate language, then back to the original language \citep{mcdonald2013anonymouth}.
The suggested translations are ranked based on their overall impact on a built-in classifier, taking into account all feature changes following the round-trip translation.
The user is tasked with addressing any grammatical and semantic issues resulting from the statistical translator.
This design aims to enhance the software's usability and permits the use of more complex feature sets than Basic-9.

Operating under the assumption that the heuristic and training corpus are available, \citet{le2015secure} discovered that the perturbed samples, modified according to Anonymouth's instructions, are reversible due to a lack of variance in feature clustering.
To address this, \citet{le2015secure} introduced randomness into the clustering, thereby making reverse engineering more difficult.
In a test using a corpus comprised of ten authors, the performance of an SVM, based on the Basic-9 feature set, fell dramatically from 86\% to 5\%.
However, these impressive results should be interpreted cautiously as the samples were rewritten by authors different from the originals, thus introducing new styles.

% % AuthorCAAT history: 
% \citet{day2016towards} -> \citet{day2016adversarial} -> \citet{faust2017adversarial} -> \citet{allred2020towards} -> \citet{packer2022towards}
Similarly, \citet{day2016towards} developed AuthorCAAT (Author Cyber Analysis \& Advisement Tool), which uses back-translation to alter style.
The intermediary languages used for back-translation were chosen as Chinese, Spanish, and English (i.e., paraphrasing within the English language itself).
AuthorCAAT employs a built-in attribution model using Manhattan distance on character uni-grams, which includes case-sensitive alphabetic characters, punctuation, and special characters.
Demonstrated with a sample of 100 bloggers, AuthorCAAT managed to decrease the inherent threat model from 55\% to 7\% while maintaining reasonable semantic preservation as evaluated by a latent semantic analysis.
Incorporating a more comprehensive set of stylistic features (than Basic-9) can produce better obfuscated samples.

Later iterations of AuthorCAAT \citep{day2016adversarial, faust2017adversarial} introduced a graphical interface to visualize writing samples from the background population, clustered and projected into a 2D space.
With these clusters, users can select a target style; AuthorCAAT then selects the most effective back-translated sentences using eight possible intermediate languages, aiming to shift the document's style towards or away from the selected.
Importantly, an automatic selection procedure, realized via a hill-climbing approach with semantic constraints, aids in reducing user fatigue when interacting with AuthorCAAT.
In their subsequent work, \citet{allred2020towards} proposed the Automated Intelligent Masking \& Information Tool (AIM-IT).
This tool replaces the back-translation component with three heuristics-based authorship masking methods reported previously in the PAN shared task.
These methods incorporate operations in accordance with a set of pre-defined rules \citep{castro2017author, mihaylova2016pan} and involve synonym substitution \citep{rahgouy2018author} (see Table~\ref{tbl: modification_based_summary} for reference).
The boundary between human-machine collaboration and fully-automatic methods becomes blurred when applications employ back-translation techniques.

\subsection{Summary}
The central question in human-machine collaboration is: how can we optimally use both human and machine abilities?
In early studies, users were burdened with extensive rewriting tasks, which did not make the best use of their effort. 
A better solution is to involve the human as a quality controller: adversarial stylometry models suggest alternative rewrites, each with its own risk of identity leak, and the user can make minor corrections to anything deemed unfit.

Improving user interface and experience design is yet another open challenge in adversarial stylometry software engineering.
A user study conducted by \citet{packer2022towards} looked into the usability of the AuthorCAAT-V software: participants found it user-friendly but prone to silent failure and struggled to follow the instructions for making suggested modifications.

Large Language Models (LLMs) trained to follow human instructions (e.g., ChatGPT) open up vast potential through prompt engineering.
Although, to the best of our knowledge, current state-of-the-art LLMs do not specifically consider the task of defending against authorship identification attacks, they have encountered various genres and styles during pretraining and have been undergone supervised fine-tuning with tasks such as summarization and simplification.
Therefore, in theory, they possess the capacity to alter an individual's style to some extent.

\section{Automatic Evasion}\label{sec: automatic}

Automatic evasion strategies offer the benefit of requiring minimal human intervention.
We classify these methods into two categories, depending on the operational space of the approach: If perturbations are introduced by directly altering the document's surface using predefined rules, potentially with the application of heuristics during or after the process, we designate the method as \emph{modification-based}. 
Conversely, a \emph{generation-based} model manipulates a document through its continuous, lower-dimensional representation prior to generating an obfuscated document.

\subsection{Modification-based Approaches}\label{sec: modification_approach}

Modification-based methods primarily operate in the symbol space, incrementally modifying a document using predefined rules. 
These operations may adhere to constraints regarding style modification, semantic preservation, and language quality.

\begin{table}[!ht]
 \caption{Summary of modification-based evasion methods. ``Constraints'' notes whether a defense considers factors other than effectiveness; ``Targeted'' denotes whether a defense is specifically tailored to a particular style.
 \label{tbl: modification_based_summary}}
 \small
  \centering
  \resizebox{\textwidth}{!}
  {
  \begin{tabular}{p{2.5cm}p{1.5cm}p{3.1cm}p{1.8cm}p{2cm}p{4cm}p{1.5cm}p{1.5cm}}
    \toprule
    Work
    & Adversary
    & Threat Models
    & Defense
    & Corpora
    & Effectiveness
    & Constraints
    & Targeted \\
    \midrule
    \citet{khosmood2010automatic}
    & Attribution
    & JGAAP 
    & Synonym replacement
    & AAAC
    & 38.5\% accuracy drop (13 authors)
    & Semantic \& grammatical
    & no \\ \cmidrule(r){1-8}
    \citet{mansoorizadeh2016author}
    & Verification
    & 44 models from PAN 2013--2015 
    & Synonym replacement
    & PAN 2016
    & 3.8\% accuracy drop
    & Semantic \& grammatical
    & No \\ \cmidrule(r){1-8}
    \citet{mihaylova2016pan}
    & Verification
    & 44 models from PAN 2013--2015 
    & Rule-based
    & PAN 2016
    & 11.0\% accuracy drop
    & -
    & No \\ \cmidrule(r){1-8}
    \citet{karadzhov2017case}
    & Verification
    & -
    & Rule-based
    & PAN 2016
    & -
    & Grammatical
    & No \\ \cmidrule(r){1-8}
    \citet{keswani2016author}
    & Verification
    & 44 models from PAN 2013--2015
    & Rule-based
    & PAN 2016
    & 10.7\% accuracy drop
    & Semantic \& grammatical
    & No \\ \cmidrule(r){1-8}
    \citet{castro2017author} 
    & Verification
    & 44 models from PAN 2013--2015 
    & Rule-based
    & PAN 2016
    & 12.8\% accuracy drop
    & -
    & No \\ \cmidrule(r){1-8}
    \citet{bakhteev2017author} 
    & Verification
    & 44 models from PAN 2013--2015 
    & Synonym substitution \& paraphrasing
    & PAN 2016
    & 7.3\% accuracy drop
    & Semantic \& grammatical
    & Yes \\ \cmidrule(r){1-8}
    \citet{rahgouy2018author}
    & Verification
    & 44 models from PAN 2013--2015 
    & Synonym substitution \& paraphrasing
    & PAN 2016
    & 9.4\% accuracy drop
    & Semantic \& grammatical
    & Yes \\ \cmidrule(r){1-8}
    \citet{kocher2018unine}
    & Verification
    & 44 models from PAN 2013--2015 
    & Rule-based
    & PAN 2016
    & 5.5\% accuracy drop
    & -
    & Yes \\ \cmidrule(r){1-8}
    Mutant-X \citep{mahmood2019mutantx} 
    & Attribution
    & MLP (Basic-9), Random Forest/SVM (Writeprints variants), \& CNN
    & Heuristic-based synonym substitution
    & EBG \& Blog
    & 64\% accuracy drop on EBG, considering 5 candidates
    & Semantic 
    & Yes \\ \cmidrule(r){1-8}
    \citet{haroon2021avengers} 
    & Attribution
    & MLP (Basic-9), SVM (Writeprints-static), \& JGAAP
    & Heuristic-based synonym substitution
    & EBG 
    & -
    & Semantic 
    & Yes \\ \cmidrule(r){1-8}
    \citet{bevendorff2019heuristic} 
    & Verification
    & Unmasking, compression model, \& 14 top models from PAN 2013--2015 
    & Heuristic-based
    & PAN 2016
    & 4.8\% accuracy drop on PAN 2013--2015 top models
    & Semantic 
    & Yes \\ \cmidrule(r){1-8}
    \citet{fernandes2018author} 
    & Attribution
    & Ruzicka similarity on character 4-grams
    & DP-based word replacement
    & Reuters \& Fan Fiction
    & ca. 30\% accuracy drop with 20 authors from Reuters
    & Semantic 
    & No \\ \cmidrule(r){1-8}
    ParChoice \citep{grondahl2020effective} 
    & Attribution
    & LSTM, CNN, \& SVM (Writeprints-static)
    & Heuristic-based paraphrasing
    & BG \& EBG
    & -
    & Semantic \& grammatical
    & Yes \\ \cmidrule(r){1-8}
    \citet{mattern2022limits} 
    & Attribution
    & BERT
    & Paraphrasing
    & IMDb \& Yelp
    & MCC drops to ca. 0.20 from 0.98, considering 10 authors from IMDb
    & Semantic \& grammatical
    & No \\ 
    \bottomrule
  \end{tabular}
  }
\end{table}

\subsubsection{Rule-based Approaches}
In rule-based approaches, a document is manipulated according to a set of predefined rules with the hope that the altered document will be difficult to associate with one's existing writings. 
These rules are often carefully crafted and intuitively useful, such as merging shorter sentences into a longer one or deleting the latter noun phrase of an appositive collocation. 
However, side constraints are either absent or weak.

% a typical rule-based study and its follow-up
In the work of \citet{mihaylova2016pan}, a proposal was made to push stylistic measures towards the population mean using a set of rules.
These targeted measures are inclusive, covering lexical aspects such as the ratio of uppercase letters, syntactic aspects (e.g., the ratio of function words), and complexity measures such as sentence length.
The transformations include sentence splitting and merging; removal and replacement of function words; spelling correction and corruption; insertion and removal of punctuation; substitution of words with their synonyms, hypernyms, or definitions; paraphrasing; and switching between uppercase and lowercase.
When a measure is lower than the corpus mean, a set of corresponding rules that can increase its value will be executed; otherwise, the opposite is done.
In addition, random syntactic noise will be injected, including switching between American and British spelling, insertion of random discourse markers, and other unification operations (e.g., regularizing all ``I've'' to ``I have'', ``4'' to ``four'', and ``+'' to ``plus'').
The method has been proven to be the most effective against 44 authorship verification approaches, penalizing performance by an average of 14\% on the PAN 2013 dataset (one of four test beds in the PAN 2016 shared task).
However, the method manipulates texts aggressively, and its results have been peer-evaluated as hard to read \citep{potthast2016author}.
While the measures are pushed towards the mean, there is no mechanism controlling the ``force'' of the push, which may result in overshooting.
In their follow-up study, \citet{karadzhov2017case} fixed this issue by keeping track of the values of the features being manipulated.
The modified method retains effectiveness with improved readability.
It is also possible to adapt the method to take up the style of a specific individual by changing the reference corpus to that individual's.
Instead of imposing a rule whenever they can push a test sample from the corresponding training data, \citet{kocher2018unine} stochastically introduce punctuation and character repetition from a distribution.

% simplification pioneer
\citet{castro2017author} also applied a set of rules to texts, which included switching contractions and their expansions contrary to one's habit, replacing words with their synonyms not found in the text, and removing explanatory apposition, fragments, discourse markers, and annotations within parentheses.
An evaluation ranked this method's effectiveness second in the PAN shared task, despite some resulting documents being difficult to understand due to inappropriate substitutions \citep{hagen2017overview}.
Notably, \citet{castro2017author} was the first to propose sentence-level simplification as a defense, an intuitive approach to ``squeeze out'' one's stylistic preference.
In their study with EBG, \citet{backes2016zoos} found that synonym replacement and spelling correction were more effective in obfuscating writing style than substituting words with common misspellings.

\paragraph{Summary}
Rule-based approaches have proven to be attractive due to their effectiveness.
This can be attributed to their potential to disrupt virtually any aspect of writing style, including lexical, syntactic, and complexity-based measures.
However, such perturbations are deterministic or exhibit minimal variability, while effective noise requires unpredictability.
Therefore, pure rule-based methods should be considered vulnerable to reversibility attacks.
For instance, an autoencoder trained with a denoising objective could reconstruct clean data from ``dirty'' data given such tight variance.
Moreover, manually crafted rules are challenging to scale or adapt to specific domains.
Despite these shortcomings, these rules can be efficiently leveraged by other approaches, such as heuristic search, or as components of the prompts sent to LLMs to inform better performance.

\subsubsection{Synonym Substitution\label{sec: syonoym_substitution}}

Replacing words with their synonyms is a strategy intended to disrupt an individual's writing style and thereby conceal their identity.\footnote{Technically, synonym substitution falls under the category of rule-based or heuristic-based approaches, depending on the strength of side constraints. It is singled out for its popularity.}
We review approaches separately depending on whether they utilize a thesaurus or embeddings as their source of synsets.

\paragraph{Thesaurus-based Approaches}

WordNet is the primary source of synsets in relevant applications. 
\citet{mansoorizadeh2016author} employed WordNet-based synonym replacement in their approach to the PAN 2016 shared task.
A replaced synonym is deemed appropriate based on two criteria: similarity, based on the depth of the two senses in the taxonomy and that of their most specific ancestor node \citep{wu1994verb}, and fluency, measured with perplexity yielded from a 4-gram language model trained on the Brown corpus. 
While its effectiveness is not as high as competing methods for this task, it retains the highest grammar and semantic ratings.
The reported effectiveness of the method should be considered conservative since it only changes one word per sentence.

% style transfer pioneer, though the method is naive
\citet{khosmood2010automatic} further consider phrase-level synonyms, which may consist of up to five consecutive words.
Preference is given to the longest match with a synonym in WordNet because these matches are less ambiguous than shorter ones and arguably introduce more noise.
The authors carefully crafted the replacement phrase by combining the substitute with the inflection of the original phrase, with the help of ConceptNet \citep{liu2004conceptnet}.
Seven out of 13 samples from the AAAC corpus (task A) were misclassified after the transformation, six of which would have otherwise been correctly classified.
The authors also found it possible to overturn a decision by intentionally adopting the style of another individual.

\paragraph{Embedding-based Approaches}
Instead of using a thesaurus, researchers have utilized word embeddings as a source, with the intuition that similar words may appear in similar locations in the embedding space.
\citet{bakhteev2017author} searched for the five nearest neighbors of a non-function word in a static word embedding (fastText) as candidates.
They then chose the sequence of synonyms that yielded the lowest perplexity according to a language model trained on Shakespeare's Sonnets corpus.
They expected the revised version to appear sufficiently Shakespearean and, therefore, significantly different from its original form.

Differential privacy approaches introduce additional noise into a word vector before searching for its nearest neighbor to enforce a certain level of privacy budget.
For instance, \citet{fernandes2017novel} introduced Laplace noise to each component of word2vec vectors before querying for the vector closest to this noisy vector, then selected a token closest to the vector.
In a subsequent study, \citet{fernandes2018author} demonstrated that a random n-dimensional Laplace noise can be obtained by selecting a random vector uniformly over the surface of an n-sphere and applying a scaling factor drawn from the Gamma distribution.
When tested on an \emph{ad hoc} Reuters corpus consisting of 20 authors, the proposed approach successfully reduced the effectiveness (from 71.1\% to 41.7\%) of similarity-based classifiers that employed character 4-grams, while a topic classifier experienced only a marginal accuracy decrease.
% leave the example as is
However, it should be noted that the resulting text is often incomprehensible due to the removal of function words and the excessive injection of noise, e.g., a transformation from ``began answered prince servants king'' to ``wildly diverging Caisse populaire Widianto Hendro Cahyono.''

\paragraph{Summary}
Synonym substitution presents a promising method for changing expressions that involve common words.
Compared to other rule-based methods, the noise injected through synonym replacements has a wider variance, as it inherits variability from both the thesaurus and the embeddings and is further enhanced through a random search from synsets.
However, function words are often left unchanged due to the difficulty of finding appropriate substitutes, and as a result, canonical syntactic patterns are only minimally disrupted when only single synonyms are replaced.
An additional concern when using static word embeddings as a source is that word embeddings, trained naively based on the distributional hypothesis, do not distinguish between antonymy and synonymy \citep{mrksic2016counter}, which may result in substantial semantic loss.

The word-level differential privacy mechanism is problematic for producing coherent documents because words are perturbed independently \citep{mattern2022limits}, thus incurring extensive noise.
Current token-level differential privacy mechanisms only deal with documents of equal length; otherwise, the required privacy budget grows linearly with the length of the output document \citep{mattern2022limits}, which violates its privacy guarantee.

\subsubsection{Heuristic Search}
Building upon predefined rules, heuristic-based methods explore the discrete space with explicitly defined constraints imposed during the search.

% classifier dependent
These constraints can be defined by considering a built-in classifier along with additional measures that track the quality of the text.
\citet{mahmood2019mutantx} proposed ``Mutant-X,'' an iterative search for optimal synonym substitutions with constraints on both effectiveness and semantic similarity.
Mutant-X iteratively replaces synonyms using ``mutation'' and ``crossover'' operations.
In each iteration, mutation replaces a portion of the words with their sentiment-aligned, nearest neighbors found in word2vec embeddings.
The crossover operation splits two mutated texts into halves at a random position and then concatenates each half with one from a different parent text.
A fitness score, considering both the confidence of an underlying classifier and the METEOR score measured between the original and transformed text, is used.
Only transformed documents scoring higher than a certain threshold are retained for the next iteration.
Following similar fitness considerations, ParChoice \citep{grondahl2020effective} combines several operations, including paraphrasing with the Paraphrase Database, synonym substitution with WordNet, grammatical transformations, and the introduction of misspellings.
Alternatively, \citet{rahgouy2018author} consider a synonym candidate as fit when it is less used by an author and can increase the Word Mover's Distance between the perturbed and the original document, apart from the similarity to the word to be substituted.

% also there are classifier independent measure; but only expose to some space (e.g., character trigrams)
Constraints can also be derived in a classifier-independent manner in certain feature spaces.
\citet{bevendorff2019heuristic} found that the Jensen-Shannon divergence of character tri-gram frequencies serves as a useful discriminator between same-author and different-author pairs.
The Jensen-Shannon divergence is a symmetric variant of Kullback-Leibler divergence, and guiding a search with this metric eliminates the requirement of an underlying classifier.
The authors used the difference in Jensen-Shannon divergence values at a specific length to guide the search because Jensen-Shannon divergence is found to be inversely correlated with log-scale text length.
The search takes into account the accumulated cost of applying operations including context-dependent word replacement and deletion, character swap and substitution, and context-free synonym and hypernym substitution.
Unlike classifier-dependent methods, the importance of this approach is arguably less influenced by the algorithm bias of the considered classifiers.
The authors suggested possible reversibility when facing a threat model with perfect knowledge of the defense \citep{bevendorff2020divergence}.

\paragraph{Summary}
By building upon predefined rules and guided by heuristics, heuristic-based methods excel at introducing high-quality noise and sufficient randomness into the text.
This makes them effective at obfuscating writing style while preserving semantic coherence.
However, due to their operation in the discrete space, the search process can be computationally expensive.
Future research could focus on enhancing the efficiency of the search process and refining the constraints used during the search.

\subsection{Generation-based Approaches}\label{sec: generation_approach}

Generation-based approaches, designed to modify writing style, employ a variety of generative models with diverse objectives and training methods.
These approaches fall into two categories: back-translation and style transfer.

\begin{table}[!ht]
 \caption{Summary of generation-based evasion methods. ``Constraints'' notes whether a defense considers factors other than effectiveness; ``Targeted'' denotes whether a defense is specifically tailored to a particular style.
 \label{tbl: generation_based_summary}}
 \small
  \centering
  \resizebox{\textwidth}{!}{
  \begin{tabular}{p{2.5cm}p{1.5cm}p{3.1cm}p{1.8cm}p{2cm}p{4cm}p{1.5cm}m{1.5cm}}
    \toprule
    Work
    & Adversary
    & Threat Models
    & Defense
    & Corpora
    & Effectiveness
    & Constraints
    & Targeted \\
    \midrule
    $\text A^4\text {NT}$ \citep{shetty2018a4nt} 
    & Attribution
    & LSTM 
    & GANs-based style transfer
    & Political speech
    & F1 drops to 0 from 1.0 considering two candidates 
    & Semantic \& grammatical
    & Yes \\ \cmidrule(r){1-8}
    ER-AE \citep{bo2020authorship} 
    & Attribution
    & Character n-gram embedding (MLP)
    & Multi-objective autoencoder 
    & Yelp
    & Accuracy drops to 9.8\% from 55.1\% considering 100 candidates 
    & Semantic \& grammatical
    & No \\ \cmidrule(r){1-8}
    % \multirow{2}{*}{\parbox{2.5cm}{\citet{emmery2018style}}}
    \multirow{4}{2.5cm}{\citet{emmery2018style}}
    & Attribution
    & Word n-gram embedding (MLP)
    & GRL-enhanced autoencoder
    & Bible
    & Accuracy drops to 24.6\% from 86.6\% considering 5 versions of the Bible 
    & Semantic \& grammatical
    & Yes \\ \cmidrule(r){2-8}
    & Attribution
    & Word n-gram embedding (MLP)
    & Machine translation
    & Bible
    & Accuracy drops to 8.0\% from 86.6\% considering 5 versions of the Bible 
    & Semantic \& grammatical
    & Yes \\ \cmidrule(r){1-8}
    % \multirow{2}{*}{\parbox{2.5cm{\citet{weggenmann2022dp_vae}}}
    \multirow{7}{2.5cm}{\citet{weggenmann2022dp_vae}}
    & Attribution
    & Word uni-/bi-gram (SVM)
    & Autoencoder with representation disentanglement
    & IMDb62 \& Yelp
    & Accuracy drops to 24\% from 90\% considering 62 candidates
    & Semantic \& grammatical
    & Yes \\ \cmidrule(r){2-8}
    & Attribution
    & Word uni-/bi-gram (SVM)
    & DP-variational autoencoder
    & IMDb62 \& Yelp
    & Accuracy drops to 14\% from 90\% considering 62 candidates
    & Semantic \& grammatical
    & Yes \\ 
    \bottomrule
  \end{tabular}}
\end{table}

\subsubsection{Back-translation}
With the initial realization that stylometric analysis could pose a threat to privacy, \citet{rao2000can} proposed a potential countermeasure: translating a document from English to another language and then back again.
Intuitively, this process of back-translation (also known as round-trip or pivot translation) has the potential to disrupt an individual's unique stylistic patterns, thereby hindering stylometric analysis.
As machine translation systems have evolved and improved, back-translation has become a standard baseline for author obfuscation research.

% commercial translator
Drawing on the findings of \citet{almishari2014fighting}, Google Translate was employed to randomly select up to nine intermediate languages.
This choice drew inspiration from earlier studies where a maximum of two intermediate languages were used, but these did not yield significant anonymization effects \citep{caliskan2012translate, brennan2012adversarial}.
It was discovered that the more intermediate languages were applied, the less likely it was to associate the test sample with an identity.
The readability of the back-translated text was generally acceptable, and a typical MTurk worker could further enhance it.
The revised review maintained its adversarial power and was evaluated as satisfactory.
Despite this, back-translation proved less effective in deceiving a stylistic classifier compared to text revised by MTurk workers, indicating potential for improvement in the statistical translation models used in the study.
% no experiment is reported on the idea of shuffling, they resorted to Moss: en-de-fr-en
\citet{keswani2016author} proposed a hypothesis: shuffling translators might introduce more randomness and consequently, improve effectiveness.

Most back-translation studies documented in the literature use statistical translation models.
However, it is widely acknowledged that contemporary attention-based models are generally more effective.
In a study by \citet{altakrori2022multifaceted}, a multilingual attention-based machine translation model \citep{fan2021beyond} was used and was found to be superior to two leading heuristic-based methods \citep{mahmood2019mutantx, bevendorff2019heuristic} in terms of effectiveness and content preservation, using the Reuters and EBG corpora.
Their findings also showed that output logits were distributed more evenly across candidates, implying a more generalized style.
This arguably ``neutral'' style is beneficial as it helps minimize misattributions to innocent candidates.

\subsubsection{Style Transfer\label{sec: style_transfer}}
Unlike back-translation, which changes style incidentally, style transfer aims to rewrite a document in a different style\footnote{The term ``style'' in style transfer is often broadly defined in a data-driven manner, encapsulating elements such as writing style, sentiment, toxicity, simplicity, humor, political slant, and even topic \citep{jin2022deep}.
% todo: reference missing
In this context, we focus on the ``writing style'' discussed in Section~\ref{sec: stylometry_style}, hence addressing writing style transfer in this section.}. 
This makes style transfer appropriate for cases where defense against authorship identification is a concern.
Early attempts at writing style transformation used either summarized guidelines \citep{hoover1999language} or predefined rules \citep{khosmood2010automatic}.
However, these methods were labor-intensive and did not scale or generalize well.
Modern style transfer, akin to back-translation, has benefited from the advances in RNN- and Transformer-based language models.

\paragraph{Approaches Using Parallel Corpora}
% it's too good to have parallel corpora
Modern language models can efficiently tackle sequence-to-sequence mapping problems using a parallel corpus, i.e., text that shares the same semantics but differs in style.
This approach has been reported to be effective in altering writing style.
For example, \citet{emmery2018style} utilized an LSTM-based encoder-decoder translation model trained on verse pairs from distinct English versions of the Bible, including the Old and New Testaments \citep{carlson2018evaluating}.
This model allows the translation of verses into specific styles derived from other Bible versions.
A unique token was included at the start of each target verse to indicate its origin.
This approach managed to generate verses that deceived a strong adversary into performing below chance levels while maintaining considerable semantic content.

Using a pretrained language model fine-tuned with a paraphrase corpus can also assist in masking personal style.
\citet{mattern2022limits} modeled the softmax function's sampling as a differential privacy mechanism and demonstrated that each token sampled requires a privacy budget of $\epsilon = 2\delta/T$ \citep{mcsherry2007mechanism}, where $\delta$ is the sensitivity and $T$ is the softmax temperature.
Thus, the total privacy budget equals $n\epsilon$, with $n$ being the number of generated tokens.
\citet{mattern2022limits} fine-tuned a pretrained GPT-2 model with sentence pairs labeled as entailment from the Natural Language Inference Corpus \citep{bowman2015large}.
When tested on obfuscated documents from ten authors in the IMDb corpus, a BERT threat model only achieved a Matthews correlation coefficient (MCC) of 0.19 ($T = 0.05$) and 0.22 ($T = 10$), compared to an MCC of 0.98 without defense.
Moreover, the perplexity was significantly lower than that of approaches using token-level differential privacy constraints.

\paragraph{Approaches Not Using Parallel Corpora}
Obtaining high-quality parallel data with sufficient linguistic variation, such as the Bible corpus, is challenging, however, and often infeasible in many domains.
As a solution, researchers employ training methods that do not depend on aligned corpora.
These methods include autoencoders, generative adversarial networks (GANs), and variational autoencoders (VAEs), possibly combined with multiple \emph{ad hoc} decoders and disentangled representations.

% methods work with non-parallel corpora: AE
\paragraph{Autoencoder-based Approaches}
Researchers have leveraged autoencoders in their studies for their ability to convert inputs into continuous representations.
\citet{bakhteev2017author} used an LSTM-based autoencoder trained to reconstruct documents from English Wikipedia.
Out of various generated sentences, those scored lowest by a language model trained on Shakespeare's texts were chosen.
This method did not have an explicit constraint for style change, but rather hoped to accidentally introduce stylistic variations during the generation sampling.

\citet{bo2020authorship} proposed the Embedding Reward Auto-Encoder (ER-AE), which uses a GRU-based autoencoder optimized for both reconstruction error and embedding rewards.
These rewards were obtained by favoring semantically similar words, as determined by cosine distance in a BERT embedding space, and by randomly sampling from the vocabulary.
When tested on a subset of the Yelp reviews corpus consisting of 100 candidates, ER-AE successfully reduced the accuracy of a character n-gram-based multilayer perceptron from 55.1\% to 9.8\%, while reportedly preserving fair semantics.

\citet{emmery2018style} enhanced an autoencoder with an auxiliary Gradient Reversal Layer (GRL), which operates on intermediate encoder embeddings.
The GRL minimizes style classification performance by acting as an identity function during forward passes and reversing gradient signs during back-propagation, leading to good representations of semantics.
They also employed a conditional decoder that targets specific styles.
This approach reduced the accuracy of a word uni- and bi-gram sentence classifier from 86.6\% to 24.6\% for five versions of the Bible.

\citet{weggenmann2022dp_vae} further investigated whether disentangling latent representations into semantic and stylistic embeddings could enhance performance of an autoencoder.
Each sub-representation was associated with two auxiliary losses.
The semantic embeddings were encouraged to predict the bag-of-words distribution of the input and penalized for authorship prediction, while the stylistic embeddings were treated in the opposite manner.
By using the average stylistic embeddings as a ``pooled'' style representing all authors during generation, a word uni- and bi-gram-based SVM model achieved an accuracy of 24\% on IMDb62, albeit with a relatively low METEOR score of 0.15.

% GANs
\paragraph{Generative Adversarial Net-based Approaches}
\citet{shetty2018a4nt} introduced $\text A^4\text {NT}$ (Adversarial Author Attribute Anonymity Neural Translation) \citep{shetty2018a4nt}, which aims at mimicking specific writing styles build up Generative Adversarial Nets (GANs).
GANs consist two components: a generator ($G$) that produces synthetic samples to imitate the target style, and a discriminator ($D$) that tries to distinguish real samples from the synthesized ones from the generator.

The corpus they used contains only two political figures: Barack Obama and Donald Trump.
To transfer style between Obama and Trump, two GANs are trained, i.e, GANs$_{o \mapsto t}$ and GANs$_{t \mapsto o}$, with stylistic, semantic, and language objectives.
Taking GANs$_{o \mapsto t}$ as an example, for the stylistic objective, a sentence from Obama ($x_{o}$) is fed into GANs$_{o \mapsto t}$'s encoder and then gets decoded using the decoder in such a way that a discriminator cannot tell the synthesized sample ($\hat x_{t}$) apart from an actual Trump sentence; the discriminator learns, at the same time, to better distinguish real Trump sentences from the synthesized ones.
The synthesized samples are approximated using Gumbel sampling.
To ensure decoded samples remain semantically relevant, the authors impose a constraint: the sentence from Trump ($x_{t}$) can be easily reconstructed using the synthesized sample ($\hat x_{t}$) and the GAN of the opposite direction (GANs$_{t \mapsto o}$).
Additionally, a language modeling loss using the target style ($x_t$) is imposed to ensure grammatical language use during generation.
The joint objective for GANs$_{o \mapsto t}$ is to minimize
\begin{align}\label{eq: a4nt}
\lambda_{\text{sty}}\min_{G_{o \mapsto t}} \max_{D_{o \mapsto t}} \mathbb{E}_{p_{x}}[\log D_{o \mapsto t}(x)] + 
\mathbb{E}_{p_{\text{z}_{o \mapsto t}}}[\log (1 - D_{o \mapsto t}(G_{o \mapsto t}(z)))] + \nonumber \\
\lambda_{\text{sem}}\mathbb{E}_{p_{\text{z}_{o \mapsto t}}}[\log G_{t \mapsto o}(z)] \nonumber + \\
\lambda_{\text{lang}}\mathbb{E}_{p_{\text{z}_{o \mapsto t}}}[\log G_{t}(z)]
\end{align}
where rows correspond to stylistic, semantic, and language loss, respectively, weighted by $\lambda$s.

$\text A^4\text {NT}$ drastically reduced an LSTM-based classifier's F1 score from 0.68 to 0.21 on a sentence level and from perfect to zero on a document level. 
However, a METEOR score of 0.29 suggests moderate semantic loss, which suggests that the impressive results come with a semantic trade-off.
The authors also considered hiding the age and gender of bloggers using a different corpus (i.e., the Blog corpus) in the same way and obtained worse effectiveness results yet much higher semantic relevance (i.e., METEOR scores of 0.69 and 0.79 for age and gender, respectively).
The observed low semantic preservation during the style transfer of political speech could be attributed to the significant domain difference between the two corpora.
A shared encoder, utilized across all three tasks, is used in conjunction with task-specific decoders during the training process.
Given that the Blog corpus is approximately 52 times larger, this shared encoder proves more effective at capturing the semantics of blog posts.
Consequently, its performance is diminished when it is applied to political speeches.

% VAE (happens to have a LDP notion)
\paragraph{Variational Autoencoder-based Approaches}
\citet{weggenmann2022dp_vae} proposed a method to anonymize personal writings using a VAE, featuring differential privacy guarantees. The aim is to derive a compact representation of input data and ensure that this representation, symbolized as $\bm{z}$, follows a specific distribution.
Given an input $x$, the encoder generates parameters for an approximate posterior distribution $q_{\phi}(z|x)$, where $\phi$ indicates the encoder parameters.
This posterior is typically treated as a multivariate Gaussian with a diagonal covariance matrix, for which the encoder supplies the mean $\mu$ and variance $\sigma^2$.
The decoder then uses sampled latent variables $z$ to derive parameters for the conditional distribution $p_{\theta}(x|z)$, with $\theta$ representing the decoder parameters.

Training a basic VAE involves maximizing the evidence lower bound, defined as
\begin{align}\label{eq: vae}
\mathbb{E}{q{\phi}(z|x)}[\log p_{\theta}(x|z)] - \text{KL}(q_{\phi}(z|x) || p_{\theta}(z))
\end{align}
where the first term encourages the decoded samples to resemble the inputs, and the second term is the Kullback-Leibler divergence between the approximate posterior and the prior on the latent variables, which acts as a regularization term that forces the latent variables to follow a standard Gaussian distribution.

The authors modified a vanilla VAE to further constrain the approximate posterior $q_{\phi}(z|x) = \mathcal{N}(z; \mu, \text{diag}(\sigma^2))$ to achieve differential privacy.
(Recall that the sensitivity of local differential privacy is the maximum of any documents, $\Delta = \max_{x, x'} \|\mu(x) - \mu(x')\|_1$.)
They applied a hyperbolic tangent transformation to the posterior, effectively bounding $\mu(x)$ within a shrunken latent space that covers 99.7\% of probability.
To avoid violating differential privacy, they also fixed the variance.
Using the IMDb62 corpus, documents anonymized with DP-VAE brought the accuracy of an SVM model down to 14\% versus the 77\% attained against documents protected by a standard VAE.

In addition to using a single latent representation $z$, the authors explored whether disentangling the latent representation into semantic embeddings $z_c$ and stylistic embeddings $z_a$ could enhance performance.
Each sub-representation was associated with two auxiliary losses.
$z_c$ was encouraged to predict the bag-of-words distribution of the input and penalized by the prediction of authorship, while $z_a$ was treated in the opposite manner.
By using an averaged $z_a$ vector as a ``pooled'' style of all authors, they were able to maintain some stylistic consistency.
However, semantic retention was low, as indicated by METEOR scores typically less than 0.2.

\subsection{Summary}

In general, generation-based approaches outperform modification-based ones, demonstrating remarkable effectiveness with both grammatical and semantic considerations properly addressed. 
However, we observe that deep learning models are often trained from scratch on \emph{ad hoc} corpora, suggesting further opportunities to utilize pre-trained LLMs as a defense. 
LLMs trained on vast corpora often perform well, possibly due to implicit regularization from the training data.
As an example, \citet{jones2022robertorroberta} fine-tuned a GPT-2 model using personal writings from Twitter and Blog datasets.
Even when fine-tuned with just 50 documents, the GPT-2 model could generate posts imitating the authorship convincingly enough to deceive a classifier.
While unconditional text generation might not directly apply to adversarial rewriting, it is worth investigating the potential of conditional text generation with LLMs.

We have also seen a surge in research aiming to generate differentially private text, optimized for utility in text classification tasks closely related to our interests.
For instance, \citet{igamberdiev2023dp_bart} introduced DP-BART, a training method that performs neuron pruning and clipping on the intermediate representation of a base BART.
This method achieves a tighter privacy budget while still producing useful text for sentiment classification tasks.

\section{Discussion\label{sec: discussion}}
Diverse strategies have been proposed to counter authorship identification attacks, but our theoretical understanding of these defenses remains underdeveloped.
These strategies exhibit a wide range, from non-professionals merely attempting to adjust their writing styles---yielding only chance-level results---to the employment of intricate methods involving deep neural networks, which can occasionally result in the production of nonsensical words.
In the following sections, we will discuss theoretical insights drawn from current studies on this subject, while highlighting the existing challenges and potential future research directions.

\subsection{Theoretical Understanding}
Defending against authorship identification attacks entails the art of crafting appropriate perturbations to the document.
Three key aspects of perturbation have been identified: randomness, direction, and magnitude.

\subsubsection{Randomness}
Perturbation extends beyond mere disruption of patterns; it involves disrupting patterns \emph{at random}. 
Early works introduced deterministic transformations, later revealed as insufficient due to susceptibility to reverse attacks \citep{le2015secure}. 
Randomness should inherently characterize a robust defense, whether it is achieved by heuristically exploring a random subset of features in the symbolic space \citep{haroon2021avengers} or employing generative models that randomly sample tokens \citep{weggenmann2022dp_vae, mattern2022limits}.

\begin{quote}
\textbf{Open Challenge I: How can we introduce randomness more effectively into defense mechanisms, while adhering to other constraints?}
\end{quote}

\subsubsection{Direction}
The resulting style of a document pertains to the ``direction'' of the perturbations.

\paragraph{Generic vs. Specific Style}

\begin{quote}
\textbf{Open Challenge II: Which type of post-transformation style is most effective: generic, specific, or somewhere in between?}
\end{quote}

Three competing theories have emerged regarding which type of style is more effective at achieving anonymization: 

\begin{enumerate}
    \item Rewriting the document to be obfuscated in a generic style. Such a supposedly ``neutral'' style, which could be practically defined with a large corpus, holds promise in masking personal markers.
    \item Adopting a specific style. This method has shown its effectiveness in corpora obtained from field studies \citep{brennan2012adversarial, wang2022reproduction} and more general-purpose corpora \citep{emmery2018style, shetty2018a4nt, weggenmann2022dp_vae}.
    \item Blending into a sub-population. \citet{backes2016zoos} found that modifying one's style to match the average of a randomly chosen sub-population is less effective than aligning it with the nearest neighbors in stylistic terms. The core idea here is to anonymize a document by blending the author's writing style into a stylistically similar sub-population.
\end{enumerate}

\paragraph{Complex vs. Simple Style}

\begin{quote}
\textbf{Open Challenge III: Which level of stylistic complexity is most promising for preserving anonymity?}
\end{quote}

\begin{enumerate}
    \item Adopting a more complex style might introduce additional signals that could overshadow person-specific styles. User studies have suggested that individuals tend to complicate their style when they are following software instructions \citep{mcdonald2012use}.
    \item Conversely, a simpler style might help reduce personal traits, leaving behind only the core content of the text. The potential of models that can explore summarization and entailment remains untapped, as suggested by \citet{potthast2016author}.
\end{enumerate}

\subsubsection{Trade-offs}

Trade-offs between utility and effectiveness arise from the amount of noise that must be injected.

\begin{quote}
\textbf{Open Challenge IV: How can we finely balance the trade-offs between anonymity and utility?}
\end{quote}

Undoubtedly, grammatically correct sentences that perfectly retain semantics are likely to facilitate successful communication.
However, finding the optimal transformation can be challenging due to the multiple constraints outlined.
If the risk of identity exposure is considered unacceptable, it may be necessary to relax grammatical and semantic constraints.
This scenario could occur when the pool of potential authors is exceptionally small, such as when only one or two individuals have access to inside information.
In such cases, a whistleblower might prioritize anonymization over message delivery.
This strategy is akin to a fail-close valve design, which automatically shuts down to prevent catastrophic outcomes due to a loss of pressure.
In our case, while communication might fail, anonymity is preserved.

In balancing utility and anonymity, we notice a convergence between differential privacy and stylometry studies. 
The former provide a theoretical framework and analytical tools that lead to a better understanding of defense mechanisms. 
Ideally, we would like a rigorous proof of privacy guarantee, which could encourage public adoption and legal admissibility. 
The latter provide historical insights into personal style and its potential linguistic implications.

Differential privacy primarily addresses information-theoretic situations, where all knowledge is accessible to the adversary. 
This assumption, however, may be excessively stringent for practical scenarios, such as in the case of a whistleblower. 
Moreover, implementations of differential privacy often involve pragmatic approximations of proposed privacy mechanisms, thereby rendering their theoretical privacy guarantees subject to empirical validation. 

Apart from the open challenges, research in the area is not immune to common problems found in the machine learning communities.
Chief among these are the infrequent reporting of empirical results in languages other than English and the lack of standard testbeds.

\subsection{Usability}
Our survey of current research reveals a concerning gap: there is a notable lack of user-friendly software designed for non-expert computer users.

\begin{quote}
\textbf{Open Challenge V: How can we craft accessible software and deliver it to users who need to anonymize their writings?}
\end{quote}

Due to the unique nature of anonymization, software distribution requires special considerations to avoid exposing user identities.
For example, delivering services through a website or APIs increases the risk of traffic analysis, making such delivery methods practically inviable.
It should be noted that, at the time of writing, none of the existing software is considered fit for practical use by non-expert computer users, primarily due to failures in software delivery and maintenance.
Furthermore, research shows that crafting accessible software for users beyond computer experts is challenging and requires dedicated user interface and experience design.
Gaining trust from users also presents a dilemma: without a thorough description of the method, it is hard to gain trust from a privacy-sensitive user, while detailing a method would make surrogate attacks feasible and eventually undermine the effectiveness of the defense.

Ultimately, we hope that in the near future we can provide whistleblowers with more detailed advice than ``Try your best to copy Cormac McCarthy.''

%Bibliography
\bibliographystyle{acl_natbib}  
\bibliography{references}

\begin{thebibliography}{126}
\expandafter\ifx\csname natexlab\endcsname\relax\def\natexlab#1{#1}\fi

\bibitem[{Abadi et~al.(2016)Abadi, Chu, Goodfellow, McMahan, Mironov, Talwar, and Zhang}]{abadi2016deep}
Martin Abadi, Andy Chu, Ian Goodfellow, H.~Brendan McMahan, Ilya Mironov, Kunal Talwar, and Li~Zhang. 2016.
\newblock \href {https://doi.org/10.1145/2976749.2978318} {Deep learning with differential privacy}.
\newblock In \emph{Proceedings of the 2016 ACM SIGSAC Conference on Computer and Communications Security}, CCS '16, pages 308--318, New York, NY, USA. Association for Computing Machinery.

\bibitem[{Abbasi and Chen(2008)}]{abbasi2008writeprints}
Ahmed Abbasi and Hsinchun Chen. 2008.
\newblock \href {https://doi.org/10.1145/1344411.1344413} {Writeprints: A stylometric approach to identity-level identification and similarity detection in cyberspace}.
\newblock \emph{ACM Transactions on Information Systems (TOIS)}, 26(2):1--29.

\bibitem[{Afroz et~al.(2012)Afroz, Brennan, and Greenstadt}]{afroz2012detecting}
Sadia Afroz, Michael Brennan, and Rachel Greenstadt. 2012.
\newblock \href {https://doi.org/10.1109/SP.2012.34} {Detecting hoaxes, frauds, and deception in writing style online}.
\newblock In \emph{2012 IEEE Symposium on Security and Privacy}, pages 461--475, USA. IEEE, IEEE Computer Society.

\bibitem[{Afroz et~al.(2014)Afroz, Islam, Stolerman, Greenstadt, and McCoy}]{afroz2014doppelganger}
Sadia Afroz, Aylin~Caliskan Islam, Ariel Stolerman, Rachel Greenstadt, and Damon McCoy. 2014.
\newblock \href {https://doi.org/10.1109/SP.2014.21} {Doppelg{\"a}nger finder: Taking stylometry to the underground}.
\newblock In \emph{2014 IEEE Symposium on Security and Privacy}, pages 212--226, USA. IEEE Computer Society.

\bibitem[{Allred et~al.(2020)Allred, Packer, Dozier, Aykent, Richardson, and King}]{allred2020towards}
Jordan Allred, Sadaira Packer, Gerry Dozier, Sarp Aykent, Alexicia Richardson, and Michael~C King. 2020.
\newblock \href {https://ieeexplore.ieee.org/document/9249682} {Towards a human-{AI} hybrid for adversarial authorship}.
\newblock In \emph{2020 SoutheastCon}, pages 1--8, USA. IEEE, IEEE.

\bibitem[{Almishari et~al.(2014)Almishari, Oguz, and Tsudik}]{almishari2014fighting}
Mishari Almishari, Ekin Oguz, and Gene Tsudik. 2014.
\newblock \href {https://doi.org/10.1145/2660460.2660486} {Fighting authorship linkability with crowdsourcing}.
\newblock In \emph{Proceedings of the Second ACM Conference on Online Social Networks}, COSN '14, pages 69--82, New York, NY, USA. Association for Computing Machinery.

\bibitem[{Altakrori et~al.(2022)Altakrori, Scialom, Fung, and Cheung}]{altakrori2022multifaceted}
Malik Altakrori, Thomas Scialom, Benjamin C.~M. Fung, and Jackie Chi~Kit Cheung. 2022.
\newblock \href {https://aclanthology.org/2022.emnlp-main.153} {A multifaceted framework to evaluate evasion, content preservation, and misattribution in authorship obfuscation techniques}.
\newblock In \emph{Proceedings of the 2022 Conference on Empirical Methods in Natural Language Processing}, pages 2391--2406, Abu Dhabi, United Arab Emirates. Association for Computational Linguistics.

\bibitem[{Baayen et~al.(2002)Baayen, van Halteren, Neijt, and Tweedie}]{baayen2002experiment}
Harald Baayen, Hans van Halteren, Anneke Neijt, and Fiona Tweedie. 2002.
\newblock \href {https://sfs.uni-tuebingen.de/~hbaayen/publications/BaayenVanHalterenNeijtTweedieJADT2002.pdf} {An experiment in authorship attribution}.
\newblock In \emph{JADT 2002: Journées Internationales d’Analyse Statistique des Données Textuelles}, volume~1, pages 69--75.

\bibitem[{Baayen et~al.(1996)Baayen, Van~Halteren, and Tweedie}]{baayen1996outside}
Harald Baayen, Hans Van~Halteren, and Fiona Tweedie. 1996.
\newblock \href {https://doi.org/doi.org/10.1093/llc/11.3.121} {Outside the cave of shadows: Using syntactic annotation to enhance authorship attribution}.
\newblock \emph{Literary and Linguistic Computing}, 11(3):121--132.

\bibitem[{Backes et~al.(2016)Backes, Berrang, and Manoharan}]{backes2016zoos}
Michael Backes, Pascal Berrang, and Praveen Manoharan. 2016.
\newblock \href {https://doi.org/10.1007/978-3-319-43005-8_3} {From zoos to safaris--from closed-world enforcement to open-world assessment of privacy}.
\newblock In \emph{Tutorial Lectures on Foundations of Security Analysis and Design VIII - Volume 9808}, pages 87--138, Berlin, Heidelberg. Springer-Verlag.

\bibitem[{Bakhteev and Khazov(2017)}]{bakhteev2017author}
Oleg Bakhteev and Andrey Khazov. 2017.
\newblock \href {https://ceur-ws.org/Vol-1866/paper_68.pdf} {Author masking using sequence-to-sequence models}.
\newblock In \emph{Working Notes of the Conference and Labs of the Evaluation Forum}, Dublin, Ireland. CEUR-WS.

\bibitem[{Barlas and Stamatatos(2020)}]{barlas2020cross}
Georgios Barlas and Efstathios Stamatatos. 2020.
\newblock \href {https://doi.org/10.1007/978-3-030-49161-1_22} {Cross-domain authorship attribution using pre-trained language models}.
\newblock In \emph{Artificial Intelligence Applications and Innovations}, pages 255--266, Cham. Springer International Publishing.

\bibitem[{Bevendorff et~al.(2019)Bevendorff, Potthast, Hagen, and Stein}]{bevendorff2019heuristic}
Janek Bevendorff, Martin Potthast, Matthias Hagen, and Benno Stein. 2019.
\newblock \href {https://doi.org/10.18653/v1/P19-1104} {Heuristic authorship obfuscation}.
\newblock In \emph{Proceedings of the 57th Annual Meeting of the Association for Computational Linguistics}, pages 1098--1108, Florence, Italy. Association for Computational Linguistics.

\bibitem[{Bevendorff et~al.(2020)Bevendorff, Wenzel, Potthast, Hagen, and Stein}]{bevendorff2020divergence}
Janek Bevendorff, Tobias Wenzel, Martin Potthast, Matthias Hagen, and Benno Stein. 2020.
\newblock \href {https://webis.de/downloads/publications/papers/bevendorff_2020a.pdf} {On divergence-based author obfuscation: An attack on the state of the art in statistical authorship verification}.
\newblock \emph{IT-Information Technology}, 62(2):99--115.

\bibitem[{Biggio and Roli(2018)}]{biggio2018wild}
Battista Biggio and Fabio Roli. 2018.
\newblock \href {https://doi.org/10.1145/3243734.3264418} {Wild patterns: Ten years after the rise of adversarial machine learning}.
\newblock In \emph{Proceedings of the 2018 ACM SIGSAC Conference on Computer and Communications Security}, CCS '18, pages 2154--2156, New York, NY, USA. Association for Computing Machinery.

\bibitem[{Bo et~al.(2021)Bo, Ding, Fung, and Iqbal}]{bo2020authorship}
Haohan Bo, Steven H.~H. Ding, Benjamin C.~M. Fung, and Farkhund Iqbal. 2021.
\newblock \href {https://doi.org/10.18653/v1/2021.naacl-main.314} {{ER}-{AE}: Differentially private text generation for authorship anonymization}.
\newblock In \emph{Proceedings of the 2021 Conference of the North American Chapter of the Association for Computational Linguistics: Human Language Technologies}, pages 3997--4007, Online. Association for Computational Linguistics.

\bibitem[{Bogoychev et~al.(2021)Bogoychev, Van~der Linde, and Heafield}]{bogoychev2021translatelocally}
Nikolay Bogoychev, Jelmer Van~der Linde, and Kenneth Heafield. 2021.
\newblock \href {https://doi.org/10.18653/v1/2021.emnlp-demo.20} {{T}ranslate{L}ocally: Blazing-fast translation running on the local {CPU}}.
\newblock In \emph{Proceedings of the 2021 Conference on Empirical Methods in Natural Language Processing: System Demonstrations}, pages 168--174, Online and Punta Cana, Dominican Republic. Association for Computational Linguistics.

\bibitem[{Bowman et~al.(2015)Bowman, Angeli, Potts, and Manning}]{bowman2015large}
Samuel~R. Bowman, Gabor Angeli, Christopher Potts, and Christopher~D. Manning. 2015.
\newblock \href {https://doi.org/10.18653/v1/D15-1075} {A large annotated corpus for learning natural language inference}.
\newblock In \emph{Proceedings of the 2015 Conference on Empirical Methods in Natural Language Processing}, pages 632--642, Lisbon, Portugal. Association for Computational Linguistics.

\bibitem[{Brennan et~al.(2012)Brennan, Afroz, and Greenstadt}]{brennan2012adversarial}
Michael Brennan, Sadia Afroz, and Rachel Greenstadt. 2012.
\newblock \href {https://doi.org/10.1145/2382448.2382450} {Adversarial stylometry: Circumventing authorship recognition to preserve privacy and anonymity}.
\newblock \emph{ACM Transactions on Information and System Security (TISSEC)}, 15(3):1--22.

\bibitem[{Brennan and Greenstadt(2009)}]{brennan2009practical}
Michael Brennan and Rachel Greenstadt. 2009.
\newblock \href {https://nyuscholars.nyu.edu/en/publications/practical-attacks-against-authorship-recognition-techniques} {Practical attacks against authorship recognition techniques}.
\newblock In \emph{21st Innovative Applications of Artificial Intelligence Conference, IAAI-09}, pages 60--65, Pasadena, California, USA. IAAI.

\bibitem[{Caliskan and Greenstadt(2012)}]{caliskan2012translate}
Aylin Caliskan and Rachel Greenstadt. 2012.
\newblock \href {https://doi.org/10.1109/ICSC.2012.46} {Translate once, translate twice, translate thrice and attribute: Identifying authors and machine translation tools in translated text}.
\newblock In \emph{2012 IEEE Sixth International Conference on Semantic Computing}, pages 121--125, Palermo, Italy. IEEE, IEEE.

\bibitem[{Carlson et~al.(2018)Carlson, Riddell, and Rockmore}]{carlson2018evaluating}
Keith Carlson, Allen Riddell, and Daniel Rockmore. 2018.
\newblock \href {https://doi.org/10.1098/rsos.171920} {Evaluating prose style transfer with the {Bible}}.
\newblock \emph{Royal Society Open Science}, 5(10):171920.

\bibitem[{Castro-Castro et~al.(2017)Castro-Castro, Bueno, and Munoz}]{castro2017author}
Daniel Castro-Castro, Reynier~Ortega Bueno, and Rafael Munoz. 2017.
\newblock \href {https://ceur-ws.org/Vol-1866/paper_170.pdf} {Author masking by sentence transformation}.
\newblock In \emph{Working Notes of the Conference and Labs of the Evaluation Forum}, Dublin, Ireland. CEUR-WS.

\bibitem[{Cer et~al.(2018)Cer, Yang, Kong, Hua, Limtiaco, St.~John, Constant, Guajardo-Cespedes, Yuan, Tar, Strope, and Kurzweil}]{cer2018universal}
Daniel Cer, Yinfei Yang, Sheng-yi Kong, Nan Hua, Nicole Limtiaco, Rhomni St.~John, Noah Constant, Mario Guajardo-Cespedes, Steve Yuan, Chris Tar, Brian Strope, and Ray Kurzweil. 2018.
\newblock \href {https://doi.org/10.18653/v1/D18-2029} {Universal sentence encoder for {E}nglish}.
\newblock In \emph{Proceedings of the 2018 Conference on Empirical Methods in Natural Language Processing: System Demonstrations}, pages 169--174, Brussels, Belgium. Association for Computational Linguistics.

\bibitem[{Chen and Dolan(2011)}]{chen2011collecting}
David Chen and William Dolan. 2011.
\newblock \href {https://aclanthology.org/P11-1020} {Collecting highly parallel data for paraphrase evaluation}.
\newblock In \emph{Proceedings of the 49th Annual Meeting of the Association for Computational Linguistics: Human Language Technologies}, pages 190--200, Portland, Oregon, USA. Association for Computational Linguistics.

\bibitem[{Clark et~al.(2019)Clark, Khandelwal, Levy, and Manning}]{clark2019does}
Kevin Clark, Urvashi Khandelwal, Omer Levy, and Christopher~D. Manning. 2019.
\newblock \href {https://doi.org/10.18653/v1/W19-4828} {What does {BERT} look at? {An} analysis of {BERT}{'}s attention}.
\newblock In \emph{Proceedings of the 2019 ACL Workshop BlackboxNLP: Analyzing and Interpreting Neural Networks for NLP}, pages 276--286, Florence, Italy. Association for Computational Linguistics.

\bibitem[{Day et~al.(2016{\natexlab{a}})Day, Brown, Thomas, Bass, and Dozier}]{day2016adversarial}
Siobahn Day, James Brown, Zachery Thomas, Lowell Bass, and Gerry Dozier. 2016{\natexlab{a}}.
\newblock \href {https://doi.org/10.1109/ICCCN.2016.7568489} {Adversarial authorship, authorwebs, and entropy-based evolutionary clustering}.
\newblock In \emph{2016 25th International Conference on Computer Communication and Networks (ICCCN)}, pages 1--6, Red Hook, NY, USA. IEEE, IEEE.

\bibitem[{Day et~al.(2016{\natexlab{b}})Day, Williams, Shelton, and Dozier}]{day2016towards}
Siobahn Day, Henry Williams, Joseph Shelton, and Gerry Dozier. 2016{\natexlab{b}}.
\newblock \href {https://ecommons.udayton.edu/maics/2016/Saturday/6/} {Towards the development of a cyber analysis \& advisement tool ({CAAT}) for mitigating de-anonymization attacks}.
\newblock In \emph{The Modern Artificial Intelligence and Cognitive Science Conference}.

\bibitem[{Domo(2020)}]{domo2020data}
Domo. 2020.
\newblock \href {https://www.domo.com/learn/infographic/data-never-sleeps-9} {Data never sleeps 9.0}.

\bibitem[{Dwork and Roth(2014)}]{dwork2014algorithmic}
Cynthia Dwork and Aaron Roth. 2014.
\newblock \href {https://doi.org/10.1561/0400000042} {The algorithmic foundations of differential privacy}.
\newblock \emph{Foundations and Trends® in Theoretical Computer Science}, 9(3–4):211--407.

\bibitem[{Eder(2015)}]{eder2015does}
Maciej Eder. 2015.
\newblock \href {https://doi.org/10.1093/llc/fqt066} {Does size matter? {Authorship} attribution, small samples, big problem}.
\newblock \emph{Digital Scholarship in the Humanities}, 30(2):167--182.

\bibitem[{Elleg{\aa}rd(1962)}]{ellegrd1962julniusletter}
Alvar Elleg{\aa}rd. 1962.
\newblock \href {https://books.google.com/books?id=gu9ZAAAAMAAJ} {\emph{A Statistical Method for Determining Authorship: The {Junius Letters}, 1769-1772}}.
\newblock Acta Universitatis Gothoburgensis. Almqvist \& Wiksell, Stockholm, Sweden.

\bibitem[{Emmery et~al.(2018)Emmery, Manjavacas~Arevalo, and Chrupa{\l}a}]{emmery2018style}
Chris Emmery, Enrique Manjavacas~Arevalo, and Grzegorz Chrupa{\l}a. 2018.
\newblock \href {https://aclanthology.org/C18-1084} {Style obfuscation by invariance}.
\newblock In \emph{Proceedings of the 27th International Conference on Computational Linguistics}, pages 984--996, Santa Fe, New Mexico, USA. Association for Computational Linguistics.

\bibitem[{Fabien et~al.(2020)Fabien, Villatoro-Tello, Motlicek, and Parida}]{fabien2020bertaa}
Ma{\"e}l Fabien, Esau Villatoro-Tello, Petr Motlicek, and Shantipriya Parida. 2020.
\newblock \href {https://aclanthology.org/2020.icon-main.16} {{B}ert{AA} : {BERT} fine-tuning for authorship attribution}.
\newblock In \emph{Proceedings of the 17th International Conference on Natural Language Processing (ICON)}, pages 127--137, Indian Institute of Technology Patna, Patna, India. NLP Association of India (NLPAI).

\bibitem[{Fan et~al.(2021)Fan, Bhosale, Schwenk, Ma, El-Kishky, Goyal, Baines, Celebi, Wenzek, Chaudhary, Goyal, Birch, Liptchinsky, Edunov, Grave, Auli, and Joulin}]{fan2021beyond}
Angela Fan, Shruti Bhosale, Holger Schwenk, Zhiyi Ma, Ahmed El-Kishky, Siddharth Goyal, Mandeep Baines, Onur Celebi, Guillaume Wenzek, Vishrav Chaudhary, Naman Goyal, Tom Birch, Vitaliy Liptchinsky, Sergey Edunov, Edouard Grave, Michael Auli, and Armand Joulin. 2021.
\newblock \href {https://dl.acm.org/doi/pdf/10.5555/3546258.3546365} {Beyond {English}-centric multilingual machine translation}.
\newblock \emph{Journal of Machine Learning Research}, 22(1).

\bibitem[{Fan(2019)}]{fan2019practical}
Liyue Fan. 2019.
\newblock \href {https://doi.org/10.1109/ICME.2019.00140} {Practical image obfuscation with provable privacy}.
\newblock In \emph{2019 IEEE International Conference on Multimedia and Expo (ICME)}, pages 784--789, Shanghai, China. IEEE.

\bibitem[{Fang et~al.(2019)Fang, Wang, Yamagishi, Echizen, Todisco, Evans, and Bonastre}]{fang2019speaker}
Fuming Fang, Xin Wang, Junichi Yamagishi, Isao Echizen, Massimiliano Todisco, Nicholas Evans, and Jean-Francois Bonastre. 2019.
\newblock \href {https://doi.org/10.21437/SSW.2019-28} {Speaker anonymization using x-vector and neural waveform models}.
\newblock In \emph{Proceedings of the 10th ISCA Workshop on Speech Synthesis (SSW 10)}, pages 155--160, Vienna, Austria.

\bibitem[{Faust et~al.(2017)Faust, Dozier, Xu, and King}]{faust2017adversarial}
Christina Faust, Gerry Dozier, Jinsheng Xu, and Michael~C King. 2017.
\newblock \href {https://ieeexplore.ieee.org/document/8285355} {Adversarial authorship, interactive evolutionary hill-climbing, and author {CAAT-III}}.
\newblock In \emph{2017 IEEE Symposium Series on Computational Intelligence (SSCI)}, pages 1--8, Honolulu, HI, USA. IEEE, IEEE.

\bibitem[{Fernandes(2017)}]{fernandes2017novel}
Natasha Fernandes. 2017.
\newblock \href {https://figshare.mq.edu.au/articles/thesis/A_novel_framework_for_author_obfuscation_using_generalised_differential_privacy/19434467/1} {\emph{A novel framework for author obfuscation using generalised differential privacy}}.
\newblock Ph.D. thesis, Macquarie University.

\bibitem[{Fernandes et~al.(2018)Fernandes, Dras, and McIver}]{fernandes2018author}
Natasha Fernandes, Mark Dras, and Annabelle McIver. 2018.
\newblock \href {https://api.semanticscholar.org/CorpusID:43942677} {Author obfuscation using generalised differential privacy}.
\newblock \emph{arXiv}, abs/1805.08866.

\bibitem[{Ge et~al.(2021)Ge, Liu, Li, Xu, Geng, Li, Tan, Sun, and Zhang}]{ge2021counterfactual}
Yingqiang Ge, Shuchang Liu, Zelong Li, Shuyuan Xu, Shijie Geng, Yunqi Li, Juntao Tan, Fei Sun, and Yongfeng Zhang. 2021.
\newblock \href {https://arxiv.org/abs/2109.01962} {Counterfactual evaluation for explainable {AI}}.
\newblock \emph{arXiv preprint arXiv:2109.01962}.

\bibitem[{Goldstein-Stewart et~al.(2009)Goldstein-Stewart, Winder, and Sabin}]{goldstein2009person}
Jade Goldstein-Stewart, Ransom Winder, and Roberta Sabin. 2009.
\newblock \href {https://aclanthology.org/E09-1039} {Person identification from text and speech genre samples}.
\newblock In \emph{Proceedings of the 12th Conference of the {E}uropean Chapter of the {ACL} ({EACL} 2009)}, pages 336--344, Athens, Greece. Association for Computational Linguistics.

\bibitem[{Golle(2006)}]{golle2006revisiting}
Philippe Golle. 2006.
\newblock \href {https://doi.org/10.1145/1179601.1179615} {Revisiting the uniqueness of simple demographics in the {US} population}.
\newblock In \emph{Proceedings of the 5th ACM Workshop on Privacy in Electronic Society}, WPES '06, pages 77--80, New York, NY, USA. Association for Computing Machinery.

\bibitem[{Gorodkin(2004)}]{gorodkin2004mcc}
Jan Gorodkin. 2004.
\newblock \href {https://doi.org/https://doi.org/10.1016/j.compbiolchem.2004.09.006} {Comparing two k-category assignments by a k-category correlation coefficient}.
\newblock \emph{Computational Biology and Chemistry}, 28(5):367--374.

\bibitem[{Grieve(2023)}]{grieve2023register}
Jack Grieve. 2023.
\newblock \href {https://doi.org/doi:10.1515/cllt-2022-0040} {Register variation explains stylometric authorship analysis}.
\newblock \emph{Corpus Linguistics and Linguistic Theory}, 19(1):47--77.

\bibitem[{Gr{\"o}ndahl and Asokan(2019)}]{grondahl2019text}
Tommi Gr{\"o}ndahl and N.~Asokan. 2019.
\newblock \href {https://doi.org/10.1145/3310331} {Text analysis in adversarial settings: Does deception leave a stylistic trace?}
\newblock \emph{ACM Computing Surveys {CSUR)}}, 52(3):1--36.

\bibitem[{Gr{\"o}ndahl and Asokan(2020)}]{grondahl2020effective}
Tommi Gr{\"o}ndahl and N.~Asokan. 2020.
\newblock \href {https://arxiv.org/pdf/1905.13464.pdf} {Effective writing style transfer via combinatorial paraphrasing}.
\newblock \emph{Proceedings on Privacy Enhancing Technologies}, 2020(4):175--195.

\bibitem[{Hagen et~al.(2017)Hagen, Potthast, and Stein}]{hagen2017overview}
Matthias Hagen, Martin Potthast, and Benno Stein. 2017.
\newblock \href {https://ceur-ws.org/Vol-1866/invited_paper_4.pdf} {Overview of the author obfuscation task at {PAN 2017}: Safety evaluation revisited}.
\newblock In \emph{Working Notes of the Conference and Labs of the Evaluation Forum}, Dublin, Ireland. CEUR-WS.

\bibitem[{Haroon et~al.(2021)Haroon, Zaffar, Srinivasan, and Shafiq}]{haroon2021avengers}
Muhammad Haroon, Fareed Zaffar, Padmini Srinivasan, and Zubair Shafiq. 2021.
\newblock \href {https://arxiv.org/pdf/2109.07028.pdf} {Avengers ensemble! {I}mproving transferability of authorship obfuscation}.
\newblock \emph{arXiv preprint arXiv:2109.07028}.

\bibitem[{Holmes(1998)}]{holmes1998evolution}
David~I. Holmes. 1998.
\newblock \href {https://doi.org/10.1093/llc/13.3.111} {The evolution of stylometry in humanities scholarship}.
\newblock \emph{Literary and Linguistic Computing}, 13(3):111--117.

\bibitem[{Hoover(1999)}]{hoover1999language}
David~L. Hoover. 1999.
\newblock \href {https://www.worldcat.org/title/39658259} {\emph{Language and style in The Inheritors}}.
\newblock University Press of America.

\bibitem[{Hu et~al.(2020)Hu, Lee, Wang, Lim, and Dai}]{hu2020deepstyle}
Zhiqiang Hu, Roy Ka-Wei Lee, Lei Wang, Ee-Peng Lim, and Bo~Dai. 2020.
\newblock \href {https://doi.org/10.1007/978-3-030-60290-1_17} {{DeepStyle}: User style embedding for authorship attribution of short texts}.
\newblock In \emph{Asia-Pacific Web (APWeb) and Web-Age Information Management (WAIM) Joint International Conference on Web and Big Data}, pages 221--229. Springer.

\bibitem[{{IARPA}(2022)}]{hiatus}
{IARPA}. 2022.
\newblock \href {https://www.iarpa.gov/newsroom/article/hiatus-identification-and-privacy-fight-it-out} {Hiatus: Identification and privacy fight it out}.

\bibitem[{Igamberdiev and Habernal(2022)}]{igamberdiev2021privacy}
Timour Igamberdiev and Ivan Habernal. 2022.
\newblock \href {https://aclanthology.org/2022.lrec-1.36} {Privacy-preserving graph convolutional networks for text classification}.
\newblock In \emph{Proceedings of the Thirteenth Language Resources and Evaluation Conference}, pages 338--350, Marseille, France. European Language Resources Association.

\bibitem[{Igamberdiev and Habernal(2023)}]{igamberdiev2023dp_bart}
Timour Igamberdiev and Ivan Habernal. 2023.
\newblock \href {https://doi.org/10.18653/v1/2023.findings-acl.874} {{DP}-{BART} for privatized text rewriting under local differential privacy}.
\newblock In \emph{Findings of the Association for Computational Linguistics: ACL 2023}, pages 13914--13934, Toronto, Canada. Association for Computational Linguistics.

\bibitem[{Jin et~al.(2022)Jin, Jin, Hu, Vechtomova, and Mihalcea}]{jin2022deep}
Di~Jin, Zhijing Jin, Zhiting Hu, Olga Vechtomova, and Rada Mihalcea. 2022.
\newblock \href {https://doi.org/10.1162/coli_a_00426} {Deep learning for text style transfer: A survey}.
\newblock \emph{Computational Linguistics}, 48(1):155--205.

\bibitem[{Johnstone(1996)}]{johnstone1996linguistic}
Barbara Johnstone. 1996.
\newblock \href {https://www.worldcat.org/title/252595827} {\emph{The linguistic individual: Self-expression in language and linguistics}}.
\newblock Oxford University Press.

\bibitem[{Jones et~al.(2022)Jones, Nurse, and Li}]{jones2022robertorroberta}
Keenan Jones, Jason R.~C. Nurse, and Shujun Li. 2022.
\newblock \href {https://cdn.aaai.org/ojs/19304/19304-28-23317-1-2-20220531.pdf} {Are you {R}obert or {R}o{BERT}a? {Deceiving} online authorship attribution models using neural text generators}.
\newblock In \emph{Proceedings of the International AAAI Conference on Web and Social Media}, volume~16, pages 429--440, Limassol, Cyprus. AAAI.

\bibitem[{Juola(2008)}]{juola2008authorship}
Patrick Juola. 2008.
\newblock \href {https://doi.org/10.1561/1500000005} {Authorship attribution}.
\newblock \emph{Foundations and Trends{\textregistered} in Information Retrieval}, 1(3):233--334.

\bibitem[{Juola(2012)}]{juola2012detecting}
Patrick Juola. 2012.
\newblock \href {https://aclanthology.org/W12-0414} {Detecting stylistic deception}.
\newblock In \emph{Proceedings of the Workshop on Computational Approaches to Deception Detection}, pages 91--96, Avignon, France. Association for Computational Linguistics.

\bibitem[{Juola and Vescovi(2010)}]{juola2010empirical}
Patrick Juola and Darren Vescovi. 2010.
\newblock \href {https://doi.org/10.1145/1866423.1866427} {Empirical evaluation of authorship obfuscation using {JGAAP}}.
\newblock In \emph{Proceedings of the 3rd ACM Workshop on Artificial Intelligence and Security}, AISec '10, pages 14--18, New York, NY, USA. Association for Computing Machinery.

\bibitem[{Kacmarcik and Gamon(2006)}]{kacmarcik2006obfuscating}
Gary Kacmarcik and Michael Gamon. 2006.
\newblock \href {https://aclanthology.org/P06-2058} {Obfuscating document stylometry to preserve author anonymity}.
\newblock In \emph{Proceedings of the {COLING}/{ACL} 2006 Main Conference Poster Sessions}, pages 444--451, Sydney, Australia. Association for Computational Linguistics.

\bibitem[{Karadzhov et~al.(2017)Karadzhov, Mihaylova, Kiprov, Georgiev, Koychev, and Nakov}]{karadzhov2017case}
Georgi Karadzhov, Tsvetomila Mihaylova, Yasen Kiprov, Georgi Georgiev, Ivan Koychev, and Preslav Nakov. 2017.
\newblock \href {https://doi.org/10.1007/978-3-319-65813-1_18} {The case for being average: A mediocrity approach to style masking and author obfuscation}.
\newblock In \emph{Experimental IR Meets Multilinguality, Multimodality, and Interaction}, pages 173--185, Cham. Springer International Publishing.

\bibitem[{Kestemont et~al.(2015)Kestemont, Moens, and Deploige}]{kestemont2015collaborative}
Mike Kestemont, Sara Moens, and Jeroen Deploige. 2015.
\newblock \href {https://doi.org/10.1093/llc/fqt063} {Collaborative authorship in the twelfth century: A stylometric study of {Hildegard} of {Bingen} and {Guibert} of {Gembloux}}.
\newblock \emph{Digital Scholarship in the Humanities}, 30(2):199--224.

\bibitem[{Keswani et~al.(2016)Keswani, Trivedi, Mehta, and Majumder}]{keswani2016author}
Yashwant Keswani, Harsh Trivedi, Parth Mehta, and Prasenjit Majumder. 2016.
\newblock \href {https://ceur-ws.org/Vol-1609/16090890.pdf} {Author masking through translation}.
\newblock \emph{Working Notes of the Conference and Labs of the Evaluation Forum}, 1609:890--894.

\bibitem[{Khosmood and Levinson(2010)}]{khosmood2010automatic}
Foaad Khosmood and Robert Levinson. 2010.
\newblock \href {https://doi.org/10.1109/ICMLA.2010.153} {Automatic synonym and phrase replacement show promise for style transformation}.
\newblock In \emph{Proceedings of the 2010 Ninth International Conference on Machine Learning and Applications}, ICMLA '10, pages 958--961, USA. IEEE Computer Society.

\bibitem[{Kocher and Savoy(2018)}]{kocher2018unine}
Mirco Kocher and Jacques Savoy. 2018.
\newblock \href {https://ceur-ws.org/Vol-2125/paper_87.pdf} {{UniNE} at {CLEF} 2018: Author masking}.
\newblock In \emph{Working Notes of the Conference and Labs of the Evaluation Forum}, Avignon, France. CEUR-WS.

\bibitem[{Koppel et~al.(2011{\natexlab{a}})Koppel, Akiva, Dershowitz, and Dershowitz}]{koppel2011unsupervised}
Moshe Koppel, Navot Akiva, Idan Dershowitz, and Nachum Dershowitz. 2011{\natexlab{a}}.
\newblock \href {https://doi.org/10.5555/2002472.2002640} {Unsupervised decomposition of a document into authorial components}.
\newblock In \emph{Proceedings of the 49th Annual Meeting of the Association for Computational Linguistics: Human Language Technologies}, pages 1356--1364, Portland, Oregon, USA. Association for Computational Linguistics.

\bibitem[{Koppel et~al.(2009)Koppel, Schler, and Argamon}]{koppel2009computational}
Moshe Koppel, Jonathan Schler, and Shlomo Argamon. 2009.
\newblock \href {https://doi.org/10.1002/asi.20961} {Computational methods in authorship attribution}.
\newblock \emph{Journal of the American Society for Information Science and Technology}, 60(1):9--26.

\bibitem[{Koppel et~al.(2011{\natexlab{b}})Koppel, Schler, and Argamon}]{koppel2011inthewild}
Moshe Koppel, Jonathan Schler, and Shlomo Argamon. 2011{\natexlab{b}}.
\newblock \href {http://www.jstor.org/stable/41486029} {Authorship attribution in the wild}.
\newblock \emph{Language Resources and Evaluation}, 45(1):83--94.

\bibitem[{Koppel et~al.(2012)Koppel, Schler, Argamon, and Winter}]{koppel2012fundamental}
Moshe Koppel, Jonathan Schler, Shlomo Argamon, and Yaron Winter. 2012.
\newblock \href {https://doi.org/10.1080/0013838X.2012.668794} {The ``fundamental problem'' of authorship attribution}.
\newblock \emph{English Studies}, 93(3):284--291.

\bibitem[{Kosseff(2022)}]{kosseff2022anonymous}
Jeff Kosseff. 2022.
\newblock \href {https://doi.org/10.7591/cornell/9781501762383.001.0001} {\emph{The united states of anonymous: How the First Amendment shaped online speech}}.
\newblock Cornell University Press, Ithaca, New York, USA.

\bibitem[{Kusner et~al.(2015)Kusner, Sun, Kolkin, and Weinberger}]{wmdkusnerb15}
Matt Kusner, Yu~Sun, Nicholas Kolkin, and Kilian Weinberger. 2015.
\newblock \href {https://proceedings.mlr.press/v37/kusnerb15.html} {From word embeddings to document distances}.
\newblock In \emph{Proceedings of the 32nd International Conference on Machine Learning}, volume~37 of \emph{Proceedings of Machine Learning Research}, pages 957--966, Lille, France. PMLR.

\bibitem[{Le et~al.(2015)Le, Safavi-Naini, and Galib}]{le2015secure}
Hoi Le, Reihaneh Safavi-Naini, and Asadullah Galib. 2015.
\newblock \href {https://doi.org/10.1007/978-3-319-24018-3_6} {Secure obfuscation of authoring style}.
\newblock In \emph{IFIP International Conference on Information Security Theory and Practice}, pages 88--103, Heraklion, Crete, Greece. Springer, Springer Cham.

\bibitem[{Li et~al.(2022)Li, Li, Ge, King, and Lyu}]{li2022text}
Jingjing Li, Zichao Li, Tao Ge, Irwin King, and Michael Lyu. 2022.
\newblock \href {https://doi.org/10.18653/v1/2022.in2writing-1.7} {Text revision by on-the-fly representation optimization}.
\newblock In \emph{Proceedings of the First Workshop on Intelligent and Interactive Writing Assistants (In2Writing 2022)}, pages 58--59, Dublin, Ireland. Association for Computational Linguistics.

\bibitem[{Liu and Singh(2004)}]{liu2004conceptnet}
Hugo Liu and Push Singh. 2004.
\newblock \href {https://doi.org/10.1023/B:BTTJ.0000047600.45421.6d} {Concept{N}et—{A} practical commonsense reasoning tool-kit}.
\newblock \emph{BT Technology Journal}, 22(4):211--226.

\bibitem[{Love(2002)}]{love2002attributing}
Harold Love. 2002.
\newblock \href {https://www.worldcat.org/title/57204668} {\emph{Attributing authorship: An introduction}}.
\newblock Cambridge University Press, New York, USA.

\bibitem[{Lundberg and Lee(2017)}]{lundberg2017unified}
Scott~M. Lundberg and Su-In Lee. 2017.
\newblock \href {https://dl.acm.org/doi/10.5555/3295222.3295230} {A unified approach to interpreting model predictions}.
\newblock In \emph{Proceedings of the 31st International Conference on Neural Information Processing Systems}, NIPS'17, page 4768–4777, Red Hook, NY, USA. Curran Associates Inc.

\bibitem[{Mahmood et~al.(2019)Mahmood, Ahmad, Shafiq, Srinivasan, and Zaffar}]{mahmood2019mutantx}
Asad Mahmood, Faizan Ahmad, Zubair Shafiq, Padmini Srinivasan, and Fareed Zaffar. 2019.
\newblock \href {https://doi.org/10.2478/popets-2019-0058} {A girl has no name: Automated authorship obfuscation using {Mutant-X}}.
\newblock \emph{Proceedings on Privacy Enhancing Technologies}, 2019(4):54--71.

\bibitem[{Mansoorizadeh et~al.(2016)Mansoorizadeh, Rahgooy, Aminiyan, and Eskandari}]{mansoorizadeh2016author}
Muharram Mansoorizadeh, Taher Rahgooy, Mohammad Aminiyan, and Mahdy Eskandari. 2016.
\newblock \href {https://ceur-ws.org/Vol-1609/16090939.pdf} {Author obfuscation using {WordNet} and language models—notebook for {PAN} at {CLEF} 2016}.
\newblock In \emph{Working Notes of the Conference and Labs of the Evaluation Forum}, pages 5--8, Évora, Portugal. CEUR-WS.

\bibitem[{Mattern et~al.(2022)Mattern, Weggenmann, and Kerschbaum}]{mattern2022limits}
Justus Mattern, Benjamin Weggenmann, and Florian Kerschbaum. 2022.
\newblock \href {https://doi.org/10.18653/v1/2022.findings-naacl.65} {The limits of word level differential privacy}.
\newblock In \emph{Findings of the Association for Computational Linguistics: NAACL 2022}, pages 867--881, Seattle, United States. Association for Computational Linguistics.

\bibitem[{Matthews(1975)}]{matthews1975mcc}
Brian~W. Matthews. 1975.
\newblock \href {https://doi.org/https://doi.org/10.1016/0005-2795(75)90109-9} {Comparison of the predicted and observed secondary structure of {T4} phage lysozyme}.
\newblock \emph{Biochimica et Biophysica Acta (BBA) - Protein Structure}, 405(2):442--451.

\bibitem[{Maximov et~al.(2020)Maximov, Elezi, and Leal-Taix{\'e}}]{maximov2020ciagan}
Maxim Maximov, Ismail Elezi, and Laura Leal-Taix{\'e}. 2020.
\newblock \href {https://doi.org/10.1109/CVPR42600.2020.00549} {{CIAGAN}: Conditional identity anonymization generative adversarial networks}.
\newblock In \emph{Proceedings of the IEEE/CVF Conference on Computer Vision and Pattern Recognition (CVPR)}, pages 5446--5455, Seattle, WA, USA. IEEE.

\bibitem[{McCallister et~al.(2010)McCallister, Grance, and Scarfone}]{mccallister2010guide}
Erika McCallister, Timothy Grance, and Karen Scarfone. 2010.
\newblock \href {https://tsapps.nist.gov/publication/get_pdf.cfm?pub_id=904990} {Guide to protecting the confidentiality of personally identifiable information {(PII)}}.

\bibitem[{McDonald et~al.(2012)McDonald, Afroz, Caliskan, Stolerman, and Greenstadt}]{mcdonald2012use}
Andrew W.~E. McDonald, Sadia Afroz, Aylin Caliskan, Ariel Stolerman, and Rachel Greenstadt. 2012.
\newblock \href {https://doi.org/10.1007/978-3-642-31680-7_16} {Use fewer instances of the letter “i”: Toward writing style anonymization}.
\newblock In \emph{Privacy Enhancing Technologies}, pages 299--318, Berlin, Heidelberg. Springer Berlin Heidelberg.

\bibitem[{McDonald et~al.(2013)McDonald, Ulman, Barrowclift, and Greenstadt}]{mcdonald2013anonymouth}
Andrew W.~E. McDonald, Jeffrey Ulman, Marc Barrowclift, and Rachel Greenstadt. 2013.
\newblock \href {https://api.semanticscholar.org/CorpusID:9301771} {Anonymouth revamped: Getting closer to stylometric anonymity}.
\newblock In \emph{PETools: Workshop on Privacy Enhancing Tools}, volume~20, Bloomington, Indiana, USA.

\bibitem[{McSherry and Talwar(2007)}]{mcsherry2007mechanism}
Frank McSherry and Kunal Talwar. 2007.
\newblock \href {http://kunaltalwar.org/papers/expmech.pdf} {Mechanism design via differential privacy}.
\newblock In \emph{48th Annual IEEE Symposium on Foundations of Computer Science (FOCS'07)}, pages 94--103. IEEE.

\bibitem[{Mendenhall(1901)}]{mendenhall1901mechanical}
Thomas~C. Mendenhall. 1901.
\newblock \href {https://en.wikisource.org/wiki/Popular_Science_Monthly/Volume_60/December_1901/A_Mechanical_Solution_of_a_Literary_Problem} {A mechanical solution of a literary problem}.
\newblock \emph{Popular Science Monthly}, 60:97--105.

\bibitem[{Mihaylova et~al.(2016)Mihaylova, Karadjov, Nakov, Kiprov, Georgiev, and Koychev}]{mihaylova2016pan}
Tsvetomila Mihaylova, Georgi Karadjov, Preslav Nakov, Yasen Kiprov, Georgi Georgiev, and Ivan Koychev. 2016.
\newblock \href {https://ceur-ws.org/Vol-1609/16090956.pdf} {{SU@ PAN’2016}: Author obfuscation—notebook for {PAN} at {CLEF 2016}}.
\newblock In \emph{Working Notes of the Conference and Labs of the Evaluation Forum}, pages 5--8.

\bibitem[{Mosteller and Wallace(1964)}]{mosteller1964inference}
Frederick Mosteller and David~L. Wallace. 1964.
\newblock \href {https://www.worldcat.org/title/181655} {\emph{Inference and disputed authorship: The {Federalist}}}.
\newblock Addison-Wesley Publishing Company, Inc.

\bibitem[{Mrk{\v{s}}i{\'c} et~al.(2016)Mrk{\v{s}}i{\'c}, {\'O}~S{\'e}aghdha, Thomson, Ga{\v{s}}i{\'c}, Rojas-Barahona, Su, Vandyke, Wen, and Young}]{mrksic2016counter}
Nikola Mrk{\v{s}}i{\'c}, Diarmuid {\'O}~S{\'e}aghdha, Blaise Thomson, Milica Ga{\v{s}}i{\'c}, Lina~M. Rojas-Barahona, Pei-Hao Su, David Vandyke, Tsung-Hsien Wen, and Steve Young. 2016.
\newblock \href {https://doi.org/10.18653/v1/N16-1018} {Counter-fitting word vectors to linguistic constraints}.
\newblock In \emph{Proceedings of the 2016 Conference of the North {A}merican Chapter of the Association for Computational Linguistics: Human Language Technologies}, pages 142--148, San Diego, California. Association for Computational Linguistics.

\bibitem[{Narayanan et~al.(2012)Narayanan, Paskov, Gong, Bethencourt, Stefanov, Shin, and Song}]{narayanan2012feasibility}
Arvind Narayanan, Hristo Paskov, Neil~Zhenqiang Gong, John Bethencourt, Emil Stefanov, Eui Chul~Richard Shin, and Dawn Song. 2012.
\newblock \href {https://doi.org/10.1109/SP.2012.46} {On the feasibility of internet-scale author identification}.
\newblock In \emph{2012 IEEE Symposium on Security and Privacy}, pages 300--314. IEEE.

\bibitem[{Neal et~al.(2017)Neal, Sundararajan, Fatima, Yan, Xiang, and Woodard}]{neal2017surveying}
Tempestt Neal, Kalaivani Sundararajan, Aneez Fatima, Yiming Yan, Yingfei Xiang, and Damon Woodard. 2017.
\newblock \href {https://doi.org/10.1145/3132039} {Surveying stylometry techniques and applications}.
\newblock \emph{ACM Computing Surveys {(CSUR)}}, 50(6):1--36.

\bibitem[{Noecker~Jr. and Ryan(2012)}]{noecker2012distractorless}
John Noecker~Jr. and Michael Ryan. 2012.
\newblock \href {http://www.lrec-conf.org/proceedings/lrec2012/pdf/238_Paper.pdf} {Distractorless authorship verification}.
\newblock In \emph{Proceedings of the Eighth International Conference on Language Resources and Evaluation ({LREC}'12)}, pages 785--789, Istanbul, Turkey. European Language Resources Association (ELRA).

\bibitem[{Ohmann(1964)}]{ohmann1964generative}
Richard Ohmann. 1964.
\newblock \href {https://www.tandfonline.com/doi/pdf/10.1080/00437956.1964.11659831} {Generative grammars and the concept of literary style}.
\newblock \emph{Word}, 20(3):423--439.

\bibitem[{Overdorf and Greenstadt(2016)}]{overdorf2016blogs}
Rebekah Overdorf and Rachel Greenstadt. 2016.
\newblock \href {https://doi.org/doi.org/10.1515/popets-2016-0021} {Blogs, {Twitter} feeds, and {Reddit} comments: Cross-domain authorship attribution}.
\newblock \emph{Proceedings on Privacy Enhancing Technologies}, 3:155--171.

\bibitem[{Packer et~al.(2022)Packer, Seals, and Dozier}]{packer2022towards}
Sadaira Packer, Cheryl Seals, and Gerry Dozier. 2022.
\newblock \href {https://doi.org/10.1007/978-3-031-05563-8_13} {Towards the improvement of {UI/UX} of a human-{AI} adversarial authorship system}.
\newblock In \emph{HCI for Cybersecurity, Privacy and Trust: 4th International Conference, HCI-CPT 2022, Held as Part of the 24th HCI International Conference, HCII 2022, Virtual Event, June 26--July 1, 2022, Proceedings}, pages 194--205. Springer.

\bibitem[{Potthast et~al.(2016)Potthast, Hagen, and Stein}]{potthast2016author}
Martin Potthast, Matthias Hagen, and Benno Stein. 2016.
\newblock \href {https://ceur-ws.org/Vol-1609/16090716.pdf} {Author obfuscation: Attacking the state of the art in authorship verification}.
\newblock In \emph{Working Notes of the Conference and Labs of the Evaluation Forum}, pages 716--749.

\bibitem[{Potthast et~al.(2018)Potthast, Schremmer, Hagen, and Stein}]{potthast2018overview}
Martin Potthast, Felix Schremmer, Matthias Hagen, and Benno Stein. 2018.
\newblock \href {https://ceur-ws.org/Vol-2125/invited_paper_16.pdf} {Overview of the author obfuscation task at {PAN} 2018: A new approach to measuring safety}.
\newblock In \emph{Working Notes of the Conference and Labs of the Evaluation Forum}.

\bibitem[{Rahgouy et~al.(2018)Rahgouy, Giglou, Rahgooy, Zeynali, and Rasouli}]{rahgouy2018author}
Mostafa Rahgouy, Hamed~Babaei Giglou, Taher Rahgooy, Hasan Zeynali, and Salar Khayat~Mirza Rasouli. 2018.
\newblock \href {https://ceur-ws.org/Vol-2125/paper_172.pdf} {Author masking directed by author’s style}.
\newblock In \emph{Working Notes of the Conference and Labs of the Evaluation Forum}.

\bibitem[{Rainie et~al.(2013)Rainie, Kiesler, Kang, Madden, Duggan, Brown, and Dabbish}]{rainie2013anonymity}
Lee Rainie, Sara Kiesler, Ruogu Kang, Mary Madden, Maeve Duggan, Stephanie Brown, and Laura Dabbish. 2013.
\newblock \href {https://www.pewinternet.org/wp-content/uploads/sites/9/media/Files/Reports/2013/PIP_AnonymityOnline_090513.pdf} {Anonymity, privacy, and security online}.
\newblock Technical report, Pew Research Center.

\bibitem[{Rao and Rohatgi(2000)}]{rao2000can}
Josyula~R. Rao and Pankaj Rohatgi. 2000.
\newblock \href {https://doi.org/10.5555/1251306.1251313} {Can pseudonymity really guarantee privacy?}
\newblock In \emph{Proceedings of the 9th Conference on USENIX Security Symposium - Volume 9}, SSYM'00, page~7, USA. USENIX Association.

\bibitem[{Ribeiro et~al.(2016)Ribeiro, Singh, and Guestrin}]{ribeiro2016should}
Marco~Tulio Ribeiro, Sameer Singh, and Carlos Guestrin. 2016.
\newblock \href {https://doi.org/10.1145/2939672.2939778} {{``Why Should I Trust You?''}: Explaining the predictions of any classifier}.
\newblock In \emph{Proceedings of the 22nd ACM SIGKDD International Conference on Knowledge Discovery and Data Mining}, KDD '16, page 1135–1144, New York, NY, USA. Association for Computing Machinery.

\bibitem[{Ribeiro et~al.(2018)Ribeiro, Singh, and Guestrin}]{ribeiro2018anchors}
Marco~Tulio Ribeiro, Sameer Singh, and Carlos Guestrin. 2018.
\newblock \href {https://dl.acm.org/doi/abs/10.5555/3504035.3504222} {Anchors: High-precision model-agnostic explanations}.
\newblock In \emph{Proceedings of the Thirty-Second AAAI Conference on Artificial Intelligence and Thirtieth Innovative Applications of Artificial Intelligence Conference and Eighth AAAI Symposium on Educational Advances in Artificial Intelligence}, AAAI'18/IAAI'18/EAAI'18. AAAI Press.

\bibitem[{Rudman(2000)}]{rudman2000non}
Joseph Rudman. 2000.
\newblock \href {https://search.informit.org/doi/10.3316/ielapa.200105720} {Non-traditional authorship attribution studies: {Ignis Fatuus} or {Rosetta Stone}?}
\newblock \emph{Bulletin (Bibliographical Society of Australia and New Zealand)}, 24(3):163--176.

\bibitem[{Shetty et~al.(2018)Shetty, Schiele, and Fritz}]{shetty2018a4nt}
Rakshith Shetty, Bernt Schiele, and Mario Fritz. 2018.
\newblock \href {https://www.usenix.org/conference/usenixsecurity18/presentation/shetty} {{A4NT}: Author attribute anonymity by adversarial training of neural machine translation}.
\newblock In \emph{27th $\{$USENIX$\}$ Security Symposium ($\{$USENIX$\}$ Security 18)}, pages 1633--1650.

\bibitem[{Somers and Tweedie(2003)}]{somers2003authorship}
Harold Somers and Fiona Tweedie. 2003.
\newblock \href {http://www.jstor.org/stable/30204914} {Authorship attribution and pastiche}.
\newblock \emph{Computers and the Humanities}, 37(4):407--429.

\bibitem[{Stamatatos(2009)}]{stamatatos2009survey}
Efstathios Stamatatos. 2009.
\newblock \href {https://doi.org/10.1002/asi.21001} {A survey of modern authorship attribution methods}.
\newblock \emph{Journal of the American Society for Information Science and Technology}, 60(3):538--556.

\bibitem[{Stamatatos(2013)}]{stamatatos2013robustness}
Efstathios Stamatatos. 2013.
\newblock \href {https://brooklynworks.brooklaw.edu/jlp/vol21/iss2/7/} {On the robustness of authorship attribution based on character n-gram features}.
\newblock \emph{Journal of Law and Policy}, 21(2):421--439.

\bibitem[{Vadhan(2017)}]{vadhan2017complexity}
Salil Vadhan. 2017.
\newblock \href {https://doi.org/10.1007/978-3-319-57048-8_7} {The complexity of differential privacy}.
\newblock In \emph{Tutorials on the foundations of cryptography: Dedicated to Oded Goldreich}, pages 347--450, Cham. Springer International Publishing.

\bibitem[{Wang et~al.(2022{\natexlab{a}})Wang, Juola, and Riddell}]{wang2022reproduction}
Haining Wang, Patrick Juola, and Allen Riddell. 2022{\natexlab{a}}.
\newblock \href {https://arxiv.org/abs/2208.07395} {Reproduction and replication of an adversarial stylometry experiment}.
\newblock \emph{arXiv preprint arXiv:2208.07395}.

\bibitem[{Wang et~al.(2021{\natexlab{a}})Wang, Riddell, and Juola}]{wang2021mode}
Haining Wang, Allen Riddell, and Patrick Juola. 2021{\natexlab{a}}.
\newblock \href {https://doi.org/10.18653/v1/2021.eacl-main.97} {Mode effects{'} challenge to authorship attribution}.
\newblock In \emph{Proceedings of the 16th Conference of the European Chapter of the Association for Computational Linguistics: Main Volume}, pages 1146--1155, Online. Association for Computational Linguistics.

\bibitem[{Wang et~al.(2021{\natexlab{b}})Wang, Xie, and Riddell}]{wang2021cross}
Haining Wang, Xin Xie, and Allen Riddell. 2021{\natexlab{b}}.
\newblock \href {https://doi.org/10.5281/zenodo.4886596} {Cross-register authorship attribution using vernacular and classical {C}hinese texts}.
\newblock In \emph{DH Benelux 2021}. Zenodo.

\bibitem[{Wang et~al.(2022{\natexlab{b}})Wang, Le, and Lee}]{wang2022upton}
Ziyao Wang, Thai Le, and Dongwon Lee. 2022{\natexlab{b}}.
\newblock \href {https://arxiv.org/pdf/2211.09717.pdf} {{UPTON}: Unattributable authorship text via data poisoning}.
\newblock \emph{arXiv preprint arXiv:2211.09717}.

\bibitem[{Warstadt et~al.(2019)Warstadt, Singh, and Bowman}]{warstadt2018neural}
Alex Warstadt, Amanpreet Singh, and Samuel~R. Bowman. 2019.
\newblock \href {https://doi.org/10.1162/tacl_a_00290} {Neural network acceptability judgments}.
\newblock \emph{Transactions of the Association for Computational Linguistics}, 7:625--641.

\bibitem[{Weggenmann et~al.(2022)Weggenmann, Rublack, Andrejczuk, Mattern, and Kerschbaum}]{weggenmann2022dp_vae}
Benjamin Weggenmann, Valentin Rublack, Michael Andrejczuk, Justus Mattern, and Florian Kerschbaum. 2022.
\newblock \href {https://doi.org/10.1145/3485447.3512232} {{DP}-{VAE}: Human-readable text anonymization for online reviews with differentially private variational autoencoders}.
\newblock In \emph{Proceedings of the ACM Web Conference 2022}, WWW '22, pages 721--731, New York, NY, USA. Association for Computing Machinery.

\bibitem[{Williams(1975)}]{williams1975mendenhalls}
C.~B. Williams. 1975.
\newblock \href {https://doi.org/10.1093/biomet/62.1.207} {{Mendenhall's studies of word-length distribution in the works of Shakespeare and Bacon}}.
\newblock \emph{Biometrika}, 62(1):207--212.

\bibitem[{Wu and Palmer(1994)}]{wu1994verb}
Zhibiao Wu and Martha Palmer. 1994.
\newblock \href {https://doi.org/10.3115/981732.981751} {Verb semantics and lexical selection}.
\newblock In \emph{32nd Annual Meeting of the Association for Computational Linguistics}, pages 133--138, Las Cruces, New Mexico, USA. Association for Computational Linguistics.

\bibitem[{Xie et~al.(2022)Xie, Wang, and Riddell}]{xie2022many}
Xin Xie, Haining Wang, and Allen Riddell. 2022.
\newblock \href {https://dh-abstracts.library.virginia.edu/works/12053} {The many voices of {Du Ying}: Revisiting the disputed writings of {Lu Xun} and {Zhou Zuoren}}.
\newblock In \emph{The Book of Abstracts of DH2022}, pages 400--404.

\bibitem[{Xu et~al.(2012)Xu, Ritter, Dolan, Grishman, and Cherry}]{xu2012paraphrasing}
Wei Xu, Alan Ritter, Bill Dolan, Ralph Grishman, and Colin Cherry. 2012.
\newblock \href {https://aclanthology.org/C12-1177} {Paraphrasing for style}.
\newblock In \emph{Proceedings of {COLING} 2012}, pages 2899--2914, Mumbai, India. The COLING 2012 Organizing Committee.

\bibitem[{Yu et~al.(2023)Yu, Buchanan, Pai, Chu, Wu, Tong, Haeffele, and Ma}]{yu2023white}
Yaodong Yu, Sam Buchanan, Druv Pai, Tianzhe Chu, Ziyang Wu, Shengbang Tong, Benjamin~D. Haeffele, and Yi~Ma. 2023.
\newblock \href {https://arxiv.org/pdf/2306.01129.pdf} {White-box transformers via sparse rate reduction}.
\newblock \emph{arXiv preprint arXiv:2306.01129}.

\bibitem[{Zhai et~al.(2022)Zhai, Rusert, Shafiq, and Srinivasan}]{zhai2022girl}
Wanyue Zhai, Jonathan Rusert, Zubair Shafiq, and Padmini Srinivasan. 2022.
\newblock \href {https://doi.org/10.18653/v1/2022.acl-long.509} {Adversarial authorship attribution for deobfuscation}.
\newblock In \emph{Proceedings of the 60th Annual Meeting of the Association for Computational Linguistics (Volume 1: Long Papers)}, pages 7372--7384, Dublin, Ireland. Association for Computational Linguistics.

\bibitem[{Zhang et~al.(2020{\natexlab{a}})Zhang, Kishore, Wu, Weinberger, and Artzi}]{zhangbertscore}
Tianyi Zhang, Varsha Kishore, Felix Wu, Kilian~Q. Weinberger, and Yoav Artzi. 2020{\natexlab{a}}.
\newblock \href {https://openreview.net/forum?id=SkeHuCVFDr} {Bertscore: Evaluating text generation with {BERT}}.
\newblock In \emph{International Conference on Learning Representations}.

\bibitem[{Zhang et~al.(2020{\natexlab{b}})Zhang, Sheng, Alhazmi, and Li}]{zhang2020adversarial}
Wei~Emma Zhang, Quan~Z. Sheng, Ahoud Alhazmi, and Chenliang Li. 2020{\natexlab{b}}.
\newblock \href {https://doi.org/10.1145/3374217} {Adversarial attacks on deep-learning models in natural language processing: A survey}.
\newblock \emph{ACM Transactions on Intelligent Systems and Technology (TIST)}, 11(3):1--41.

\bibitem[{Zhao and Chen(2022)}]{zhao2022survey}
Ying Zhao and Jinjun Chen. 2022.
\newblock \href {https://doi.org/10.1145/3490237} {A survey on differential privacy for unstructured data content}.
\newblock \emph{ACM Computing Surveys}, 54(10s).

\bibitem[{Zhu and Jurgens(2021)}]{zhu2021idiosyncratic}
Jian Zhu and David Jurgens. 2021.
\newblock \href {https://doi.org/10.18653/v1/2021.emnlp-main.25} {Idiosyncratic but not arbitrary: Learning idiolects in online registers reveals distinctive yet consistent individual styles}.
\newblock In \emph{Proceedings of the 2021 Conference on Empirical Methods in Natural Language Processing}, pages 279--297, Online and Punta Cana, Dominican Republic. Association for Computational Linguistics.

\end{thebibliography}

\end{document}